\documentclass{article}
\usepackage{tocloft}
\usepackage{multirow}
\usepackage{makecell}
\usepackage{wrapfig}
\usepackage[most]{tcolorbox}
\tcbset{
  promptbox/.style={
    colback=gray!5,
    colframe=gray!30,
    boxrule=0.8pt,
    arc=2mm,
    fonttitle=\bfseries,
    top=1mm,
    bottom=1mm,
    left=2mm,
    right=2mm,
    boxsep=1mm,
    sharp corners=south,
    listing only,
    listing options={basicstyle=\ttfamily\small, breaklines=true}
  }
}
\newcolumntype{Y}{>{\raggedright\arraybackslash}X}
\newcolumntype{P}[1]{>{\centering\arraybackslash}p{#1}} % 水平和垂直方向居中
\newcolumntype{M}[1]{>{\centering\arraybackslash}m{#1}} % 水平居中 + 垂直居中（需要 array + m）
\usepackage[preprint]{neurips_2025}

\usepackage{tabularx}
\usepackage[utf8]{inputenc} % allow utf-8 input
\usepackage[T1]{fontenc}    % use 8-bit T1 fonts
\usepackage{amsmath} % for \text in math mode
\usepackage{hyperref}       % hyperlinks
\usepackage{url}            % simple URL typesetting
\usepackage{booktabs}       % professional-quality tables
\usepackage{colortbl}       % color for table rows
\usepackage{amsfonts}       % blackboard math symbols
\usepackage{nicefrac}       % compact symbols for 1/2, etc.
\usepackage{microtype}      % microtypography
\usepackage{xcolor}         % colors
\usepackage{graphicx}
\usepackage{subcaption}  % 推荐使用subcaption而非subfigure
\usepackage{caption}

\definecolor{darkgreen}{rgb}{0.0, 0.5, 0.0}
\title{Spatial Understanding from Videos:\\ Structured Prompts Meet Simulation Data}

\author{
\textbf{Haoyu Zhang}$^{1,2}$, \quad 
\textbf{Meng Liu}$^{3*}$, \quad
\textbf{Zaijing Li}$^{1,2}$, \quad
\textbf{Haokun Wen}$^{1}$, \\
\textbf{Weili Guan}$^{1}$\textbf{,} \quad
\textbf{Yaowei Wang}$^{1,2}$\textbf{,} \quad
\textbf{Liqiang Nie}$^{1}$\thanks{Corresponding author.} \\[1ex]
$^1$Harbin Institute of Technology (Shenzhen) \quad
$^2$Pengcheng Laboratory \quad
\\$^3$Shandong Jianzhu University \\[1ex]
\texttt{https://github.com/Hyu-Zhang/SpatialMind}
}

\begin{document}

\maketitle

\begin{abstract}
Visual-spatial understanding, the ability to infer object relationships and layouts from visual input, is fundamental to downstream tasks such as robotic navigation and embodied interaction. 
However, existing methods face spatial uncertainty and data scarcity, limiting the 3D spatial reasoning capability of pre-trained vision-language models (VLMs).
To address these challenges, we present a unified framework for enhancing 3D spatial reasoning in pre-trained VLMs without modifying their architecture.
This framework combines SpatialMind, a structured prompting strategy that decomposes complex scenes and questions into interpretable reasoning steps, with ScanForgeQA, a scalable question-answering dataset built from diverse 3D simulation scenes through an automated construction process designed for fine-tuning. 
Extensive experiments across multiple benchmarks demonstrate the individual and combined effectiveness of our prompting and fine-tuning strategies, and yield insights that may inspire future research on visual-spatial understanding.
\end{abstract}

\section{Introduction}
Visual-spatial understanding, the ability to infer spatial relationships and the layout of objects from visual input, is a core component of human perception~\cite{yang2024thinking,zhang2023uncovering,jin2024llava}. From a single image, human observers can intuitively estimate distances, relative sizes, and even infer occluded structures. As intelligent systems become increasingly embedded in real-world applications such as autonomous driving~\cite{zhang2023attribute,tiandrivevlm}, robotic navigation~\cite{zhang2021multimodal,driess2023palm}, and augmented reality~\cite{chandrasegaran2024hourvideo,zhang2025exo2ego,feng2024objectnlq,feng2025object,chu2025technicalreportego4dlongterm}, it becomes crucial to endow models with similar spatial reasoning capabilities for robust perception and interaction.

Unfortunately, a single image is inherently limited in capturing the complexity of real-world 3D scenes, constraining its utility in practical scenarios~\cite{chen2024spatialvlm,cheng2025spatialrgpt,cai2024spatialbot}. To address this, point clouds have become a mainstream representation for 3D scene understanding due to their ability to encode rich geometric information~\cite{chen2024ll3da,hong20233d}. Yet, generating high-quality point clouds typically requires expensive sensors and incurs significant computational overhead, limiting scalability and accessibility. 

%This motivates the exploration of vision-only solutions that rely solely on scanning videos (or multi-view images) of scenes, offering a more human-like and accessible approach to spatial reasoning~\cite{qi2025gpt4scene}.

% Despite growing interest, current approaches to visual-spatial understanding face persistent challenges in both accuracy and generalization. Image-based methods are often limited by their reliance on 2D representations, struggling to model 3D structures and exhibiting sensitivity to viewpoint variations. Conversely, 3D point cloud-based methods~\cite{chen2024ll3da,hong20233d} provide richer geometric information but require costly data acquisition and intensive computation, limiting their scalability. These trade-offs have motivated the exploration of vision-only solutions that rely solely on RGB images or videos, offering a more human-like and accessible approach to spatial reasoning~\cite{qi2025gpt4scene}.

These limitations motivate the pursuit of vision-only solutions that operate on scanning videos or multi-view images of scenes. Such approaches offer a more human-like and scalable pathway to spatial understanding~\cite{qi2025gpt4scene,pmlr-v235-zhang24aj}. However, performing 3D spatial reasoning from scanning videos presents two significant challenges:
\textbf{(1) Spatial Uncertainty.}
In the absence of explicit depth information, models must infer 3D structure from inherently limited 2D observations. This process is further complicated by occlusions, perspective distortions, and texture ambiguities, all of which introduce significant spatial uncertainty. Effectively addressing this challenge demands multi-step logical reasoning across frames to reconstruct coherent spatial layouts. 
\textbf{(2) Data Scarcity.}
Existing datasets for this task are limited in both scale and diversity, restricting the ability of vision-language models (VLMs) to acquire robust spatial knowledge and perceptual capabilities. Moreover, these datasets involve scans of real-world scenes, which leads to poor scalability.
This highlights the need for scalable and extensible data sources to support effective spatial reasoning in VLMs.

% However, spatial reasoning from egocentric videos, remains inherently difficult. Without explicit depth signals, models must contend with occlusions, perspective distortion, and texture ambiguity—factors that introduce significant spatial uncertainty. Addressing such tasks requires not only perceptual understanding but also commonsense reasoning and multi-step logical inference. For instance, identifying the nearest object involves estimating relative distances and making a comparative judgment.

To address these challenges, we propose a dual approach for enhancing 3D spatial reasoning in pre-trained VLMs, without modifying their underlying architecture. First, we introduce \textbf{SpatialMind}, a structured Chain-of-Thought (CoT) prompting strategy that guides VLMs through step-by-step reasoning over spatial relationships. Second, we present \textbf{ScanForgeQA}, a large-scale synthetic question-answering (QA) dataset constructed from diverse 3D simulation scenes using an automated generation pipeline. Fine-tuning VLMs on this dataset equips them with spatial commonsense knowledge, significantly improving their generalization to unseen spatial layouts. We have validated our approach through extensive experiments across multiple benchmarks. Results demonstrate the individual and combined effectiveness of our prompting and fine-tuning strategies, and yield insights that may inspire future research on visual-spatial understanding.

Our contributions are summarized as follows:
\begin{itemize}
    \item We introduce SpatialMind, a spatial prompting strategy that decomposes spatial reasoning into structured steps, enabling pre-trained VLMs to perform multi-step inference over spatial relationships from visual input alone.
    \item We develop a scalable dataset generation pipeline to construct ScanForgeQA, a synthetic spatial question-answering dataset that enables VLMs to acquire spatial commonsense through fine-tuning.
    \item Experimental results validate the effectiveness and generalizability of both SpatialMind and ScanForgeQA, with their combination achieving further gains and providing valuable insights for future research.
\end{itemize}

\section{Related Work}
\textbf{2D Image Spatial Understanding} focuses on modeling spatial relationships among objects within the 2D image. Most existing models are trained on 2D images paired with textual descriptions, which offer limited cues about 3D structure. Consequently, their capacity for spatial reasoning remains constrained. To mitigate this,  several approaches, such as SpatialVLM~\cite{chen2024spatialvlm}, SpatialRGPT~\cite{cheng2025spatialrgpt}, and SpatialBot~\cite{cai2024spatialbot}, have been proposed. These methods enhance the spatial understanding by fine-tuning models on datasets specifically designed for spatially grounded QA tasks. 
To enable more comprehensive evaluation, recent studies~\cite{du2024embspatial,zhang2024sphere} have introduced hierarchical benchmarks that assess models across varying levels of spatial reasoning complexity.
Parallel efforts have explored more explicit forms of spatial interaction~\cite{maspatialpin,yu2024rag}.
For example, point-based methods~\cite{yuanrobopoint,song2024robospatial} interpret spatial instructions by predicting specific target points. Building on this trend, SpatialCoT~\cite{liu2025spatialcot} proposes a two-stage strategy that aligns multimodal inputs with spatial coordinates and incorporates CoT reasoning to better address complex embodied tasks.
Despite these advancements, model performance often degrades in complex real-world 3D environments, highlighting the limitations of 2D-based approaches in representing complex 3D scenes.  
%their performance degrades significantly when transferred to real-world  environments. This decline is primarily attributed to the inherent limitations of independent 2D images in representing complex 3D scenes.

\textbf{3D Indoor Spatial Understanding} focuses on enabling intelligent agents to identify object positions and infer their spatial relationships within enclosed environments, thereby supporting both object manipulation and interactive scene comprehension~\cite{zhang2024hcqa,zhang2025hcqa}. Early 3D models are trained on standard indoor datasets~\cite{baruch1arkitscenes,chang2017matterport3d,dai2017scannet,deitke2022️,mao2022multiscan,ramakrishnan2habitat,yeshwanth2023scannet++} using point clouds to facilitate downstream tasks like 3D object detection and instance segmentation~\cite{vu2022softgroup,wu2024point,nguyen2024open3dis,rozenberszki2024unscene3d} and primarily focus on object-level geometry and appearance features~\cite{guo2023point,qi2024shapellm,liu2024lion,xu2024pointllm}. More recent work extends this focus to complex indoor scenes, emphasizing inter-object spatial relationships and holistic scene-level understanding. To address challenges such as geometric complexity and annotation sparsity, many of these models employ cross-modal strategies that combine point cloud data with auxiliary multi-view 2D images~\cite{chen2024ll3da,hong20233d,man2024lexicon3d}.
Inspired by the way humans perceive spatial layouts through vision alone, emerging research~\cite{yang2024thinking,liao2025improved,qi2025gpt4scene} has begun to explore purely vision-based approaches to 3D spatial understanding. These methods rely solely on visual inputs, such as scanning videos, without requiring explicit 3D priors like point clouds. This line of work offers a more practical and scalable alternative for real-world deployment. In this context, we further investigate whether purely vision-based inputs can provide a more effective solution for indoor scene understanding.

\section{SpatialMind Prompting Strategy}
\begin{figure}
  \centering
  \includegraphics[width=\linewidth]{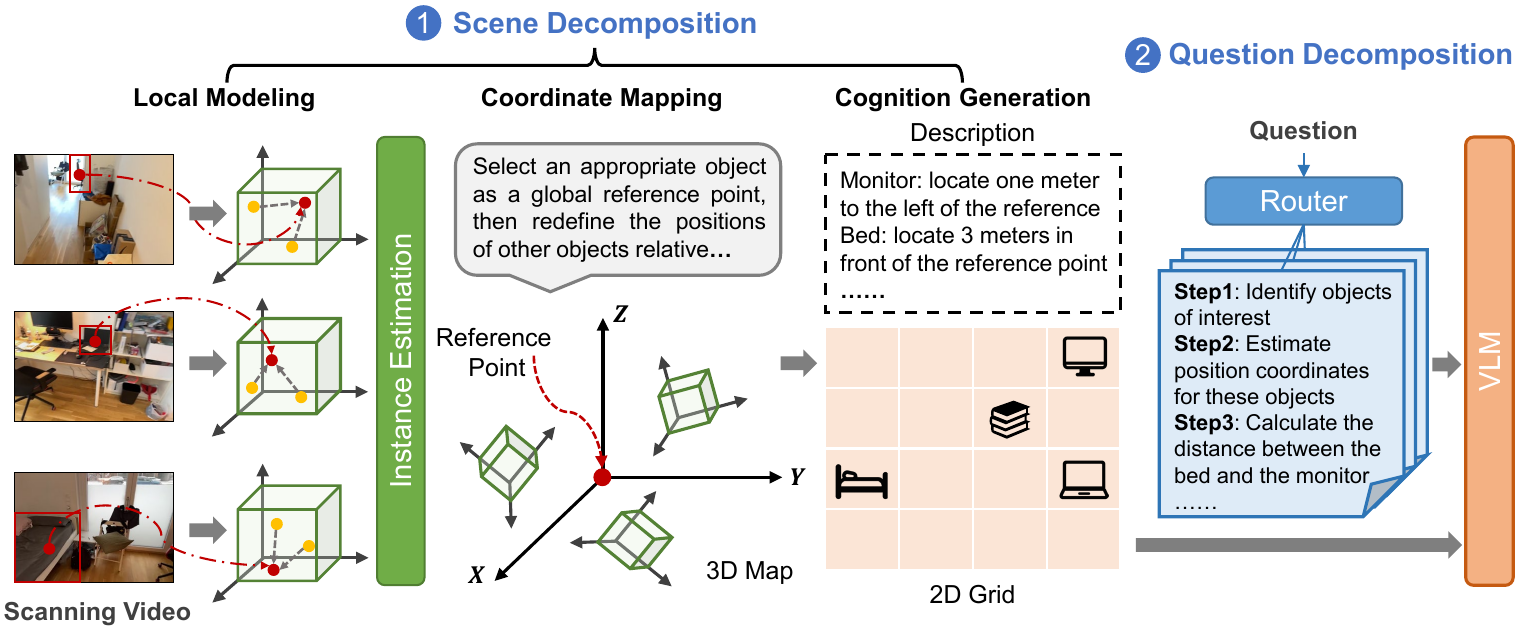}
  \caption{Illustration of our SpatailMind prompting strategy.}
  \label{fig:cot}
  \vspace{-2ex}
\end{figure}

As shown in Figure~\ref{fig:cot}, our SpatialMind prompting strategy consists of two main components: \textbf{1) Scene Decomposition}, where the 3D scene depicted in the video is transformed into multiple different representations; and \textbf{2) Question Decomposition}, in which the question is broken down into a sequence of fine-grained reasoning steps. Further details can be found in \textbf{Appendix~\ref{app_cot}}.

\subsection{Scene Decomposition}
The scene decomposition process includes three sequential steps: local modeling, coordinate mapping, and cognition generation.

\textbf{Local Modeling}. The first step processes scanning video frames to extract object instances and their relative spatial configurations within localized coordinate systems. To handle scene complexity and reduce the search space, we leverage GPT-4o\footnote{\url{https://openai.com/index/hello-gpt-4o/}.} to identify all objects mentioned across the questions associated with a given scene, using them as candidate targets. For each frame \(i\), we prompt VLMs to detect a subset of objects \( \{c_{ij}\} \) from the candidate targets and estimate their positions \( \mathbf{p}_{ij}^{\text{local}} \in \mathbb{R}^3 \). These positions are defined relative to a randomly selected reference object (i.e., origin) within the same frame, forming a local 3D map:
\begin{equation}
    \mathcal{L}_i = \left\{ (c_{ij}, \mathbf{p}_{ij}^{\text{local}}) \mid j = 1, \dots, n_i \right\},
\end{equation}
where $n_i$ denotes the number of objects in frame $i$. Because each video frame captures only a limited field of view, the same object may appear across multiple frames from different perspectives. Thus, this step focuses on accurate per-frame object detection and spatial localization, laying the foundation for subsequent alignment in a global coordinate system.

% This step captures the egocentric spatial layout of objects across frames, effectively segmenting and localizing instances under various viewpoints.
%Importantly, since each video frame reflects a narrow field of view, the same object may appear in multiple frames under different perspectives. Local modeling therefore emphasizes accurate per-frame instance detection and spatial anchoring, which later enables alignment in the global coordinate system.

\textbf{Coordinate Mapping}. To integrate spatial information across video frames, this step transforms all locally detected object positions into a unified global coordinate system. The global origin is defined by selecting the reference object in the first frame.
To estimate motion between frames, we prompt the VLM to infer the relative rotation and translation between adjacent frames. These relative transformations are accumulated sequentially to compute each frame’s transformation $\mathbf{T}_i$ with respect to the global coordinate system:
\begin{equation}
    \mathbf{T}_i = \prod_{k=1}^{i}
\begin{bmatrix}
\mathbf{R}_{k,k-1} & \mathbf{t}_{k,k-1} \\
\mathbf{0} & 1
\end{bmatrix},
\end{equation}
where $\mathbf{R}_{k,k-1}$ and $\mathbf{t}_{k,k-1}$ denote the relative rotation and translation from frame $k-1$ to frame $k$, respectively. This accumulated approach provides more stable and accurate alignment than directly estimating each frame’s absolute pose. Using these transformations, each object’s local coordinates are converted into global coordinates via homogeneous transformation: 
\begin{equation}
    \begin{bmatrix}
\mathbf{p}_{ij}^{\text{global}} \\
1
\end{bmatrix}
=
\mathbf{T}_i \cdot
\begin{bmatrix}
\mathbf{p}_{ij}^{\text{local}} \\
1
\end{bmatrix},
\end{equation}

% \[
% \mathbf{p}_{ij}^{\text{global}} = \mathbf{R}_i \cdot \mathbf{p}_{ij}^{\text{local}} + \mathbf{t}_i,
% \]
where $\mathbf{p}_{ij}^{\text{global}}$ denotes the global coordinates of the object $j$ in the frame $i$. This step ensures that all detected objects across frames are positioned consistently within the same 3D space. Since objects may appear in multiple frames under different perspectives, we merge duplicate detections based on spatial proximity and semantic consistency via prompting. The result is a global 3D map of the scene:
% After transformation, all objects are projected into a shared global 3D map. Redundant detections—i.e., multiple views of the same object—are merged based on spatial proximity and category agreement. The result is a global object set:
\begin{equation}
    \mathcal{G} = \left\{ (c_k, \mathbf{p}_k^{\text{global}}) \right\}_{k=1}^{N},
\end{equation}
where $N$ is the total number of all object instances in the entire scene. This map serves as a unified spatial abstraction that captures the overall layout from egocentric scanning videos.

\textbf{Cognition Generation}. Beyond constructing a 3D map, we explore two additional formats for representing scene structure: a 2D spatial grid and natural language descriptions. 
% With a complete global map, the system proceeds to generate a structured spatial representation for reasoning.
We define a regular 2D grid over the global scene, typically aligned with the \(XY\)-plane. Each grid cell corresponds to a fixed real-world area (e.g., 1 meter per cell, denoted by cell size \(s\) ). Each object \(c_k\) is mapped to a discrete grid location $(i_k, j_k)$:
\begin{equation}
    (i_k, j_k) = \left( \left\lfloor \frac{x_k}{s} \right\rfloor, \left\lfloor \frac{y_k}{s} \right\rfloor \right),
\end{equation}
where \( (x_k, y_k) \) are the horizontal components of the object’s global position \( \mathbf{p}_k^{\text{global}} \). In parallel, we generate natural language descriptions of object locations relative to a designated reference point. Using prompting, the model produces statements such as \{``monitor'': ``locate 1 meter to the left of the reference point''\}. These descriptions serve as a human-interpretable form of spatial cognition, bridging visual perception and symbolic reasoning. 

% These spatially grounded outputs, combined with egocentric video, are finally passed to VLMs to support downstream tasks like spatial question answering. This cognitive generation step enables abstract, language-level reasoning grounded in concrete visual observations and structured spatial priors.

\subsection{Question Decomposition}
Different types of spatial questions require distinct reasoning strategies. To accommodate this diversity, we first categorize questions into several types (e.g., object size, relative distance, and relative direction). For each category, we design a dedicated reasoning procedure using GPT-4o, followed by human verification to ensure correctness and interpretability. For instance, consider a question from the ``relative distance'' category: \textit{Among the refrigerator, window, and microwave, which object is closest to the door?} The reasoning process for this type follows four structured steps: 1) Identify all mentioned objects, 2) Estimate the spatial coordinates of all relevant objects, 3) Compute the pairwise distances between the door and each candidate object,  and 4) Select the object with the minimum distance as the answer. During inference, the system correspondingly selects the appropriate reasoning procedure based on the identified question type.

To perform 3D spatial reasoning, we feed the VLMs with the input scanning video, one form of scene representation (e.g., 3D map, 2D grid, or textual position descriptions), the question, and the corresponding step-by-step reasoning plan. To assess which scene representation format is most interpretable for VLMs, we have conducted comparative experiments, as shown in Figure~\ref{fig:format_compare}.

\section{ScanForgeQA Dataset Construction}
\begin{figure}
  \centering
  \includegraphics[width=\linewidth]{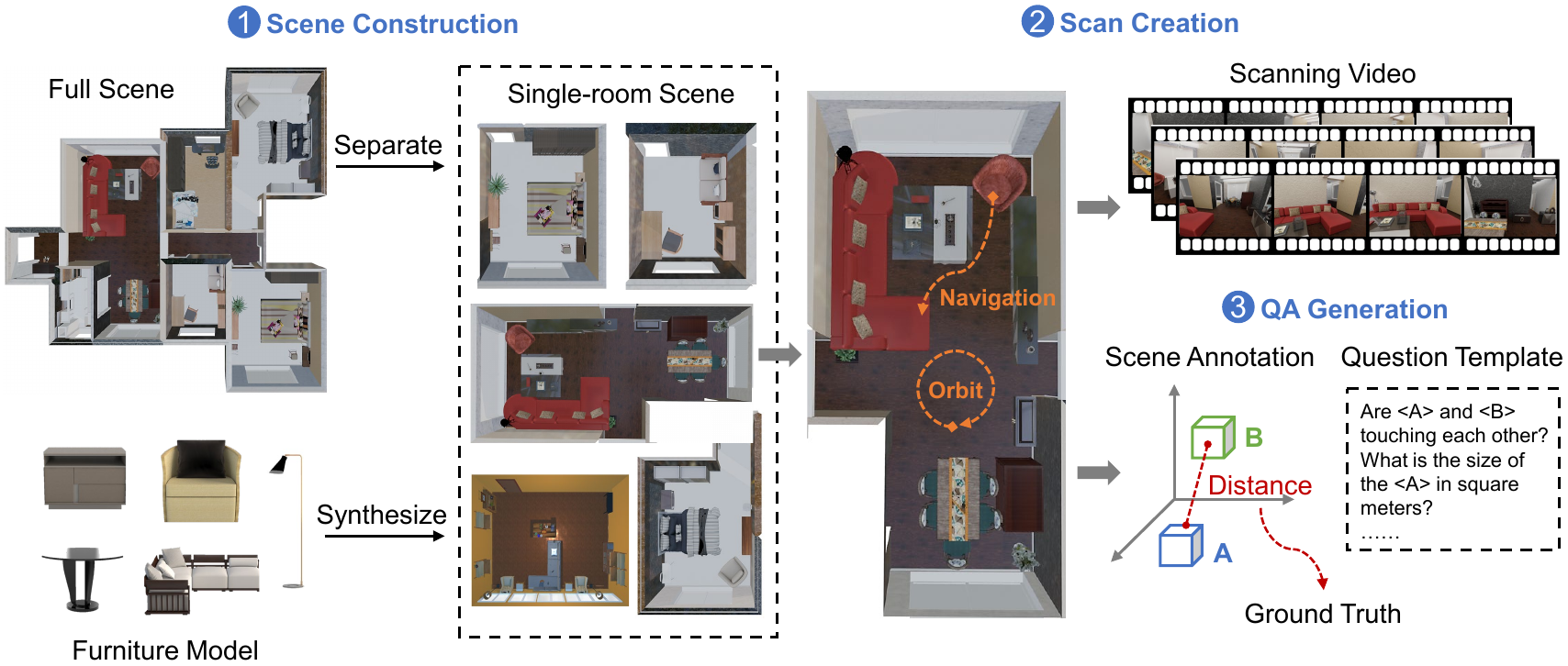}
  \caption{The pipeline of ScanForgeQA data construction.}
  \label{fig:data_gen}
  \vspace{-2ex}
\end{figure}

The construction of the \textbf{ScanForgeQA} dataset involves a three-stage pipeline, illustrated in Figure~\ref{fig:data_gen}. These stages are: \textbf{1) Scene Construction}, where single-room 3D environments are created; \textbf{2) Scan Creation}, in which egocentric videos are simulated by scanning through the constructed scenes; and \textbf{3) QA Generation}, where textual question-answering pairs are automatically generated based on object annotations and the spatial layout of each scene.

\subsection{Scene Construction}
To ensure diversity and richness in single-room scene collection, we adopt two parallel strategies:

% 可添加数据集示例
\textbf{Separation}. We modify existing scene datasets to leverage available resources effectively. Specifically, we utilize the 3D-FRONT dataset~\cite{fu20213d}, which contains 6,813 multi-room scenes furnished with diverse 3D objects and annotated with detailed layout semantics and high-quality textures. Since our focus is on single-room environments, we disassemble each multi-room scene into individual rooms. For each scene, we isolate and load one room at a time, along with its corresponding ceiling and walls, and save it as an independent instance. This disassembly process yields 44,427 single-room scenes. We further filter out uncommon room types (e.g., garage, auditorium) and those lacking sufficient object content (e.g., aisle, stairwell). The final dataset consists of 34,116 single-room scenes across six common categories: bedroom, kitchen, bathroom, living room, dining room, and storage room.
% And the type label of each room is also recorded according to the scene annotations. 

\textbf{Synthesis}. To introduce additional diversity and originality, we synthesize novel room layouts using a LLM-guided generation approach. Specifically, we adopt HoloDeck~\cite{yang2024holodeck}, a 3D generation framework that leverages LLMs to parse natural language prompts, retrieve matching assets from large-scale 3D object repositories such as Objaverse~\cite{deitke2023objaverse}, and optimize their spatial arrangement to form semantically meaningful scenes. To drive the generation process, we first use GPT-4o to create diverse textual descriptions for various room types. For example, a bedroom may be described as: \textit{``A bedroom with a bed, window, armchair, and wardrobe''}. We define eight room categories, including two additional types—office and store—and generate 20 distinct descriptions for each. These prompts are fed into HoloDeck to produce corresponding room layouts, with human verification to ensure spatial plausibility and realism. This synthesis process yields 160 additional single-room scenes.

% % 问题类型分布
% \begin{table*}[h]
% \caption{Data distribution.}
% \centering
% \begin{tabular}{ccccccccc}
% \toprule
%  & \makecell{Living\\Room} & \makecell{Kitchen} & \makecell{Bathroom} & \makecell{Bedroom} & \makecell{Office} & \makecell{Restaurant} & \makecell{Storage\\Room} & \makecell{Store} \\ 
% \midrule
% Number &  &  &  &  &  &  &  &  \\ 
% Scan &  &  &  &  &  &  &  &  \\
% QA &  &  &  &  &  &  &  &  \\
%     \bottomrule
% \end{tabular}
% \label{tab:example}
% \end{table*}
% 抖动

\subsection{Scan Creation}
To simulate egocentric scanning videos from the constructed single-room scenes, we implement a scanning procedure using the Unity engine\footnote{\url{https://unity.com/}.}. Each scene is scanned using two complementary strategies designed to emulate natural human visual exploration:

\textbf{Orbit Scan}. We define a circular trajectory centered in the room at a height of approximately 1.5 meters, corresponding to typical adult eye level. The circle’s diameter is set to two-thirds of the shorter side of the room. The camera is randomly initialized at a point on this path and moves along the circle either clockwise or counterclockwise. An image is captured every 5 degrees of rotation, resulting in 72 frames per orbit scan. This strategy provides a comprehensive 360-degree panoramic view of the scene.

\textbf{Navigation Scan}. To simulate movement through the environment, we label navigable ground regions based on object categories and generate a navigation mesh using the \textit{NavMesh Baking} API. We randomly select two objects as the navigation start and end points and compute the shortest path between them on the mesh. Among the candidate paths, the two longest are chosen for scanning to achieve a more complete coverage of the scene. For each path, the camera first performs a 360-degree rotation at the starting point, capturing an image every 12 degrees (30 images total). It then traverses the path toward the destination, during which 12 frames are uniformly sampled. Upon arrival, another 360-degree rotation is performed, again capturing 30 images. In total, 72 frames are recorded per path. Due to the limited size of indoor environments, rotational movement yields more visual variation than translation; hence, fewer frames are captured during motion.

% 答案生成示意图
\subsection{QA Generation}
To generate diverse supervised fine-tuning (SFT) data and enhance the 3D spatial reasoning capabilities of existing VLMs, we define three categories of question types: attribute estimation, spatial reasoning, and hypothesis analysis.  These categories encompass both quantitative and qualitative dimensions, and cover both open-set and closed-set scenarios. Below, we describe each category in detail, along with the methodology for deriving corresponding ground-truth answers.

\textbf{Attribute Estimation}. This type focuses on static properties of objects and scenes, such as \textit{object count} (``How many chairs are in the room?''), \textit{object size} (``What is the length of the longest side of the refrigerator in meters?''), \textit{room size} (``What is the size of this room in square meters?''), and \textit{room type} (``Based on the object layout, what is the most likely type of room (e.g., kitchen)?''). Ground-truth answers for these questions are directly derived from 3D scene annotations and object metadata provided in the dataset. 

% 或者第一张图修改为答案判断示意图
\begin{wrapfigure}{r}{0.5\textwidth}  % r=右侧图，l=左侧图；0.4宽度可调整
    \centering
    \includegraphics[width=0.48\textwidth]{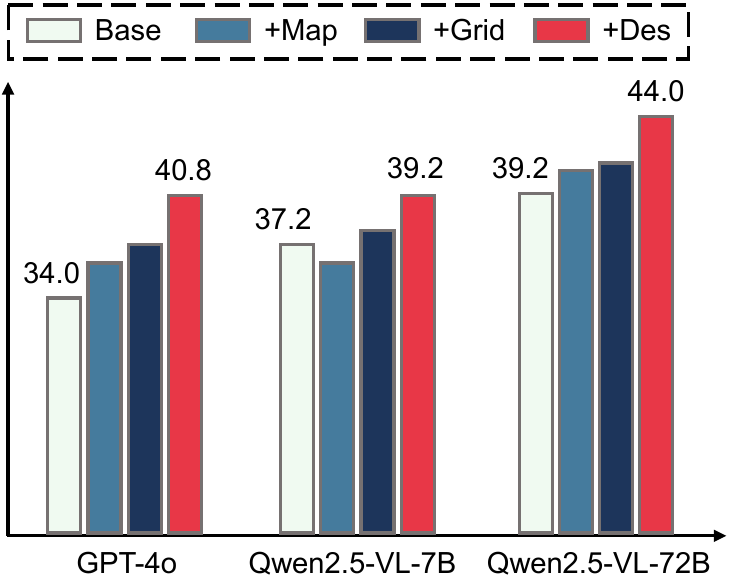}
    \caption{Effects of different scene expression.}
    \label{fig:format_compare}
\end{wrapfigure}

\textbf{Spatial Reasoning}. This category targets inter-object spatial relationships, requiring models to infer positional and geometric properties such as distance, orientation, and contact. Representative question types include: \textit{relative distance} (``Which of these objects (refrigerator, couch, ceiling light) is closest to the TV?''), \textit{absolute distance} (``What is the distance between the couch and the table in meters?''), \textit{relative direction} (``If I am standing by sofa and facing the table, which side is the trash can on?''), and \textit{contact relationship} (``Is there a gap between the bed and the headboard?''). For distance-related questions, we compute Euclidean distances between object centroids in the global 3D coordinate space. For contact relationships, object dimensions are also considered to determine physical adjacency. To resolve relative direction, we define an object’s front as the side oriented toward the room center. Angular sectors are divided clockwise into four directional categories: right (45°–135°), back (135°–225°), left (225°–315°), and front (315°–45°). For example, an object located at 80° relative to the reference point is classified as being on the right. 

\textbf{Hypothesis Analysis}. This category introduces conditional reasoning under hypothetical scenarios, often requiring geometric and commonsense inference.  A typical example is \textit{operation feasibility} (``Considering only object dimensions, is it feasible to place the television on the table?''). Feasibility is determined by comparing object dimensions. For stacking, the movable object’s length and width must be smaller than those of the supporting surface. For embedding (e.g., fitting an item into a drawer), the object’s height must also fall within the bounds of the specific container’s volume. 

The full ScanForgeQA dataset includes 34,276 single-room scenes, 103K simulated video scans, and 925K question-answering pairs for training. Leveraging synthetic environments allows scalable and controlled data generation across diverse spatial scenarios. Additional implementation details are provided in the \textbf{Appendix~\ref{app_data}}.

\begin{table}
  \caption{Performance comparison on VSI-Bench. \textcolor{red}{\raisebox{0.2ex}{$\dagger$}} indicates results on VSI-Bench (tiny) set.}
  \centering
  \setlength{\tabcolsep}{4pt} % 缩小列间距
  \begin{tabular}{lcccccccccc}
    \toprule
    \textbf{Method} & 
    \makecell{\textbf{Obj.} \\ \textbf{Count}} & 
    \makecell{\textbf{Abs.} \\ \textbf{Dist.}} & 
    \makecell{\textbf{Obj.} \\ \textbf{Size}} & 
    \makecell{\textbf{Room} \\ \textbf{Size}} & 
    \makecell{\textbf{Rel.} \\ \textbf{Dist.}} & 
    \makecell{\textbf{Rel.} \\ \textbf{Dir.}} & 
    \makecell{\textbf{Route} \\ \textbf{Plan}} & 
    \makecell{\textbf{Appr.} \\ \textbf{Order}} & 
    \textbf{Avg}&
    $\Delta$\\

    % \textbf{Method} & \textbf{Obj. Count} & \textbf{Abs. Dist.} & \textbf{Obj. Size} & \textbf{Room Size} & \textbf{Rel. Dist.} & \textbf{Rel. Dir.}$^{\star}$ & \textbf{Room Plan} & \textbf{Appr. Order} & \textbf{Avg.}\\
    \midrule
    \multicolumn{11}{c}{\textbf{Close-source}}\\
    \midrule
     Human Level\textcolor{red}{\raisebox{0.2ex}{$\dagger$}}& 94.3 & 47.0 & 60.4 & 45.9& 94.7& 95.8& 95.8 &100.0& 79.2&-\\
     Gemini-1.5 Pro\textcolor{red}{\raisebox{0.2ex}{$\dagger$}}& 49.6& 28.8& 58.6& 49.4& 46.0& 48.1&  42.0&  68.0& 48.8&-\\
     Gemini-1.5 Pro &  56.2 & 30.9 & 64.1& 43.6& 51.3& 46.3& 36.0& 34.6& 45.4&-\\
     \rowcolor{gray!20}+SpatialMind&63.9&51.8&70.2&47.3&56.3&45.9&42.6&44.3&52.8&\textbf{{\color{darkgreen} $\uparrow$ 7.4\%}}\\
     GPT-4o&46.2 &5.3 &43.8& 38.2 &37.0& 41.3& 31.5& 28.5&34.0&-\\
     \rowcolor{gray!20}+SpatialMind&40.0&27.1&62.7&40.9&41.0&39.6&37.1&38.5&40.8&\textbf{{\color{darkgreen} $\uparrow$ 6.8\%}}\\
     \midrule
    \multicolumn{11}{c}{\textbf{Open-source}}\\
    \midrule
     % DeepSeek& \\
     % +SpatialMind&\\
     InternVL2-8B & 23.1 &28.7& 48.2 &39.8& 36.7 &30.7& 29.9 &39.6& 34.6&-\\
     +SpatialMind&35.8&28.9&49.7&44.4&37.2&34.8&35.1&45.5&38.9&\textbf{{\color{darkgreen} $\uparrow$ 4.3\%}}\\
     
     +ScanForgeQA&45.3&33.4&54.8&45.0&41.1&36.1&33.4&43.0&41.5&\textbf{{\color{darkgreen} $\uparrow$ 6.9\%}}\\
          \rowcolor{gray!20}+Both&47.0&32.8&53.2&46.6&39.8&36.8&37.9&47.5&42.7&\textbf{{\color{darkgreen} $\uparrow$ 8.1\%}}\\
     InternVL2-40B & 34.9& 26.9& 46.5& 31.8 &42.1& 32.2& 34.0 &39.6& 36.0&-\\
     +SpatialMind&36.4&30.0&49.1&41.8&43.8&36.1&35.6&50.0&40.4&\textbf{{\color{darkgreen} $\uparrow$ 4.4\%}}\\
     +ScanForgeQA&51.0&29.2&52.7&38.1&47.2&36.4&35.9&47.6&42.3&\textbf{{\color{darkgreen} $\uparrow$ 6.3\%}}\\
          \rowcolor{gray!20}+Both&52.2&30.5&54.4&41.0&50.5&37.0&40.2&50.3&44.5&\textbf{{\color{darkgreen} $\uparrow$ 8.5\%}}\\
    \midrule
     % InternVL2.5-8B\\
     % InternVL2.5-38B\\
    %  LLaVA-NV-7B & 48.5 &14.0 &47.8& 24.2 &43.5& 42.4 &34.0 &30.6& 35.6&-\\
    %  +SpatialMind&49.0&22.6&47.9&24.6&41.3&43.0&37.1&29.8&36.9&\textbf{{\color{darkgreen} $\uparrow$ 1.3\%}}\\
    %       +ScanForgeQA&\\
    %            \rowcolor{gray!20}+Both&\\
    %  LLaVA-NV-72B & 48.9 & 22.8 &57.4 &35.3& 42.4 &36.7 &35.0& 48.6&40.9&-\\
    %  +SpatialMind&52.5&24.5&60.2&37.7&42.7&39.5&37.3&51.0&43.2&\textbf{{\color{darkgreen} $\uparrow$ 2.3\%}}\\
    %  +ScanForgeQA&\\
    %       \rowcolor{gray!20}+Both&\\
    % \midrule
     % LLaVA-OneVision-72B & 43.5& 23.9 &57.6& 37.5 &42.5& 39.9& 32.5& 44.6& 40.2\\
     % VideoLLaMA3-7B&36.8&22.2&34.7&24.9&44.6&41.7&36.1&28.8&33.7&-\\
     % +SpatialMind&\\     +ScanForgeQA&\\
     %      \rowcolor{gray!20}+Both&\\
    Qwen2.5-VL-7B&40.3&22.2&50.1&38.9&38.0&40.7&31.4&35.9&37.2&-\\
     +SpatialMind& 45.1&25.2&52.1&41.4&38.7&41.6&34.7&34.5&39.2&\textbf{{\color{darkgreen} $\uparrow$ 2.0\%}}\\
     +ScanForgeQA&53.2&30.5&56.8&44.9&42.3&44.0&37.3&37.7&43.3&\textbf{{\color{darkgreen} $\uparrow$ 6.1\%}}\\
     \rowcolor{gray!20}+Both&55.0&29.5&57.3&44.0&43.5&44.3&38.3&39.2&43.9&\textbf{{\color{darkgreen} $\uparrow$ 6.7\%}}\\
        Qwen2.5-VL-72B&37.9& 28.6&57.4&49.8&45.5&38.4&20.6&35.4&39.2&-\\
     +SpatialMind&42.3&32.0&61.7&53.8&48.2&43.9&30.4&39.3&44.0&\textbf{{\color{darkgreen} $\uparrow$ 4.8\%}}\\
     +ScanForgeQA&45.2&32.7&63.3&52.4&50.1&41.7&32.8&40.2&44.8&\textbf{{\color{darkgreen} $\uparrow$ 5.6\%}}\\
     \rowcolor{gray!20}+Both&48.6&34.4&68.9&54.7&53.4&43.9&30.1&42.7&47.1&\textbf{{\color{darkgreen} $\uparrow$ 7.9\%}}\\
    \bottomrule
  \end{tabular}
    \vspace{-2ex}
  \label{tab:vsi-bench}
\end{table}

\section{Experiments}
The experimental settings (including benchmarks, baselines, etc.) and more experimental results can be found in the \textbf{Appendix~\ref{app_setting}} and\textbf{~\ref{app_result}}.

\begin{table}[t]
\centering
\begin{minipage}{0.6\textwidth}
    \centering
\caption{Performance comparison on the EM-EQA subset of OpenEQA and the validation set of ScanQA and SQA3D.}
\begin{tabular}{lccc}
\toprule
\multirow{2}{*}{\textbf{Method}} & \textbf{OpenEQA}& \textbf{ScanQA}& \textbf{SQA3D}\\

&Acc/Score & BLEU-1&EM-1\\
\midrule
% Human Level & 87.7 & 85.1 & 86.8 \\
% GPT-4 & 32.5 & 35.5& 33.5\\
% GPT-4V & 55.3\\
% GPT-4o \\
%      \rowcolor{gray!20}+SpatialMind&\\
%      \midrule
Qwen2.5-VL-7B& 50.1/3.1&32.5&17.2\\
     +SpatialMind&53.7/3.2&33.1&19.8\\
     +ScanForgeQA&56.2/3.3&34.8&23.3\\
     \rowcolor{gray!20}+Both&58.6/3.5&37.9&24.5\\
Qwen2.5-VL-72B&53.8/3.2&35.4&34.8\\
     +SpatialMind&55.7/3.2&38.0&39.2\\
     +ScanForgeQA&59.1/3.4&42.5&43.0\\
     \rowcolor{gray!20}+Both&60.4/3.4&44.1&46.3\\
\bottomrule
\end{tabular}
    \label{tab:openeqa}\end{minipage}
\hfill
\begin{minipage}{0.38\textwidth}
    \centering
\caption{Effects of different fine-tuning data and prompting strategy.}
\begin{tabular}{lcc}
\toprule
\textbf{Method} & \makecell{\textbf{Room} \\ \textbf{Size}}&\textbf{Avg} \\
\midrule
Qwen2.5-VL-7B&38.9&37.2\\
+SQA3D&38.8&38.9\\  
+ScanQA&38.5& 39.1\\  
\rowcolor{gray!20}+ScanForgeQA&44.9&43.3\\  
Qwen2.5-VL-72B&49.8&39.2\\
+CoT-Question & 50.6& 41.3 \\  
+CoT-Scene&52.1&42.7\\  
\rowcolor{gray!20}+SpatialMind&53.8&44.0\\  
\bottomrule
\end{tabular}
    \label{tab:ablation}
\end{minipage}
\vspace{-3ex}
\end{table}

\subsection{Performace Comparison}
We investigated the following five key questions to assess our approach:

% We reported the results of our method and fine-tuning data across different model architectures, parameter scales, and evaluation datasets, as shown in Tables~\ref{tab:vsi-bench} and~\ref{tab:openeqa}. Detailed experimental analysis is presented below:

\textbf{Q1: Which scene representation format is most interpretable by VLMs?} Figure~\ref{fig:format_compare}  presents a performance comparison across different representation formats: no additional spatial context (Base), inclusion of a 3D map (+Map), a 2D grid (+Grid), and object-centric textual descriptions (+Des). Across all models, a consistent trend emerges: the +Des variant outperforms others, followed by +Grid, while +Map yields the least improvement. These results suggest that current VLMs are more adept at interpreting one-dimensional textual descriptions than high-dimensional structured spatial formats. Consequently, we adopted the textual description format in subsequent experiments as the default scene representation. 

\textbf{How do SpatialMind and ScanForgeQA impact VLM performance?}
As shown in Table~\ref{tab:vsi-bench}, we progressively applied the SpatialMind prompting strategy and the ScanForgeQA fine-tuning data across a range of VLMs, varying in architectures, parameter size, and openness (including both open- and closed-source models). The results reveal three key findings:
1) Both SpatialMind prompting and ScanForgeQA fine-tuning consistently improve visual-spatial understanding across models. This includes large-scale proprietary models such as Gemini-1.5 Pro~\footnote{\url{https://deepmind.google/technologies/gemini/pro/}.} and GPT-4o, demonstrating the effectiveness and generalizability of our approaches. 
2) Model size affects the relative benefit of prompting versus fine-tuning. Larger models (e.g., 72B) benefit more from prompting, which enhances their reasoning capabilities, while smaller models (e.g., 7B) show greater improvements through fine-tuning. For instance, Qwen2.5-VL-7B gains 6.1\% from fine-tuning, compared to only 2.0\% from prompting.
3) Humans and VLMs exhibit complementary strengths. Human participants excel in qualitative tasks (e.g., achieving 100\% accuracy on the \textit{Appearance Order} task) but perform poorly on precise quantitative estimations (e.g., \textit{Object Size}). In contrast, VLMs show strong quantitative reasoning ability and, in some cases, even surpass human-level performance. This contrast underscores the potential of VLMs to complement human perception in spatial tasks.

% Under the same architecture, models with larger parameter sizes exhibit more substantial performance gains compared to their smaller counterparts—for instance, InternVL-40B shows a greater improvement than InternVL-8B (4.3 vs. XXX). \textbf{Fine-tuning Setting}. For open-source models, we fine-tune them using the constructed SRSFT data. The results similarly demonstrate the superiority of our approach, further validating the effectiveness of the proposed SRSFT dataset in enhancing the spatial understanding capabilities of VLMs. Moreover, we observe that supervised fine-tuning is particularly beneficial for smaller-scale models, in contrast to the SpatialMind prompting strategy.

\textbf{Can combining prompting and fine-tuning yield further gains?}
To assess whether SpatialMind and ScanForgeQA provide complementary benefits, we applied the SpatialMind prompting strategy to models that have already been fine-tuned on the ScanForgeQA dataset. The results, reported in the ``+Both'' row of Table~\ref{tab:vsi-bench}, show consistent performance improvements across all evaluated models. These findings confirm that the two approaches are complementary.
%These results demonstrate that applying the SpatialMind prompting strategy to fine-tuned VLMs leads to further gains, confirming the complementary effect of the two approaches.

\begin{figure}[t]
  \centering
  \begin{minipage}[t]{0.48\textwidth}
    \centering
    \includegraphics[width=\linewidth]{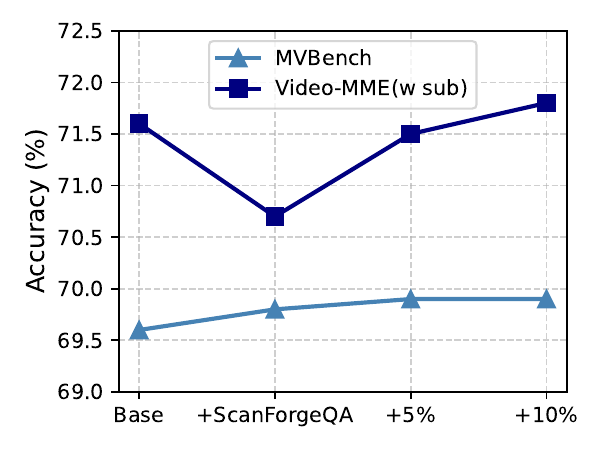}
    \caption{Performance of Qwen2.5-VL-7B on MVBench and Video-MME.}
    \label{fig:mvbench}
  \end{minipage}
  \hfill
  \begin{minipage}[t]{0.48\textwidth}
    \centering
    \includegraphics[width=\linewidth]{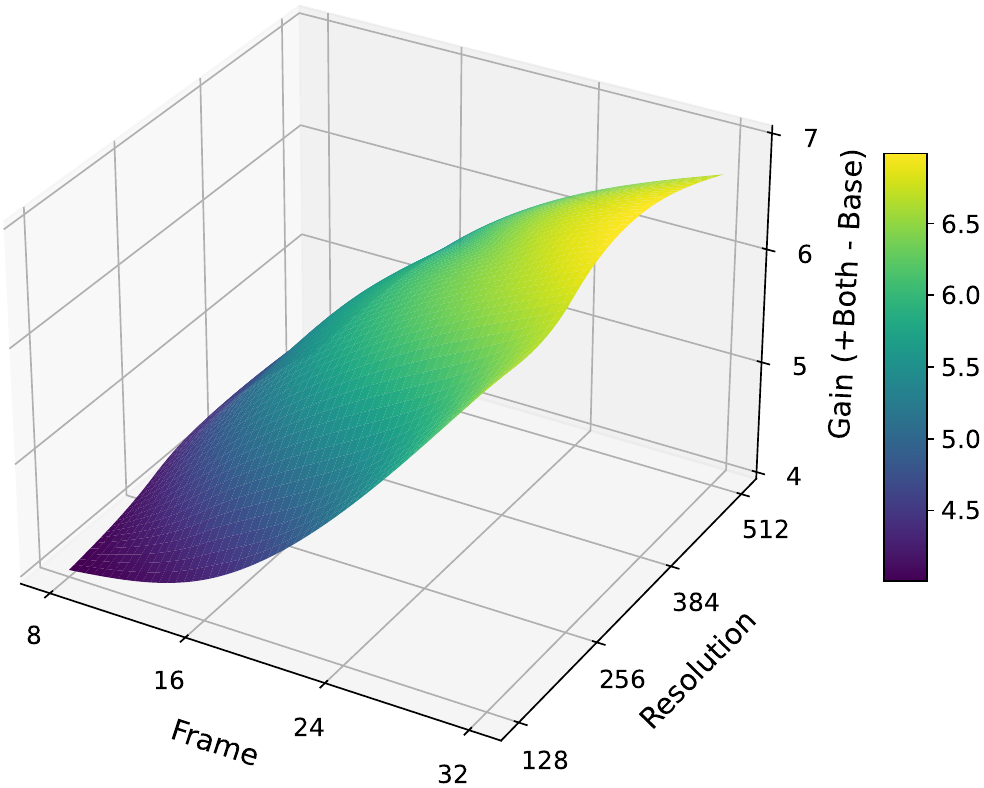}
    \caption{Ablation study of Qwen2.5-VL-7B under varying numbers of frames and resolution.}
    \label{fig:ablation}
  \end{minipage}
  \vspace{-3ex}
\end{figure}

% 或者第一张图修改为答案判断示意图
% \begin{wrapfigure}{r}{0.5\textwidth}  % r=右侧图，l=左侧图；0.4宽度可调整
%     \centering
%     \includegraphics[width=0.48\textwidth]{img/gain.pdf}
%     \caption{Ablation study on Qwen2.5-VL-7B with different frames and resolution.}
%     \label{fig:gain}
% \end{wrapfigure}

\textbf{Does the improvement generalize to other spatial benchmarks?} To assess the generalizability of our framework, we conducted evaluations on multiple benchmarks, including OpenEQA~\cite{majumdar2024openeqa}, ScanQA~\cite{azuma2022scanqa}, and SQA3D~\cite{masqa3d}. 
As shown in Table~\ref{tab:openeqa}, both SpatialMind prompting and ScanForgeQA fine-tuning lead to consistent performance gains across all benchmarks. These results validate the robustness of our approach and confirm its applicability across diverse spatial tasks and datasets. 
%provides further evidence of the effectiveness of our prompting strategy and fine-tuning data, as reflected in consistently superior performance across a range of spatial reasoning benchmarks. These results substantiate the robustness and generalizability of our framework across diverse data sources and task categories.

\textbf{Does fine-tuning affect performance on other tasks?} To investigate whether enhancing visual-spatial capabilities via fine-tuning adversely impacts a model's general performance,  we conducted evaluations on MVBench~\cite{li2024mvbench} and Video-MME~\cite{fu2024video}, two broad multi-task video benchmarks. As shown in Figure~\ref{fig:mvbench}, fine-tuning with ScanForgeQA slightly improves performance on MVBench but leads to a marginal drop on Video-MME. This difference likely stems from MVBench containing spatial reasoning tasks, while Video-MME focuses more on event comprehension. To mitigate this trade-off, we further experimented with mixed fine-tuning, combining a small proportion (5\% and 10\%) of traditional data from ShareGPT4Video~\cite{chen2024sharegpt4video} with ScanForgeQA. Results show that this strategy achieves improved performance, surpassing the original Qwen2.5-VL-7B baseline, suggesting that spatial fine-tuning can be harmonized with broader capabilities through data balancing. 
%The results are promising, with mixed fine-tuning achieving enhanced performance that surpasses the original Qwen2.5-VL-7B model.

% \begin{figure}[h]
% \centering
% \begin{minipage}[t]{0.38\textwidth}
%   \centering
%   \includegraphics[width=\textwidth]{img/mvbench.pdf} % 替换为你的图像路径
%   \caption{The results on MVBench.}
%   \label{fig:example}
% \end{minipage}
% \hfill
% \begin{minipage}[t]{0.48\textwidth}
%   \centering
%   \captionof{table}{Ablation study on Qwen2.5-VL-7B with different frames and resolution.}
%   \begin{tabular}{ccc}
%     \toprule
%     \textbf{Frame} & \textbf{Resolution} & \textbf{Avg} \\
%     \midrule
%     \multirow{3}{*}{8} & 128 \\
%      & 256 \\
%      & 512 \\
%      \midrule
%     16 & 512 &33.1\\
%     \midrule
% \rowcolor{gray!20}    32 & 512 &37.2\\
%     \bottomrule
%   \end{tabular}
%   \label{tab:example}
% \end{minipage}
% \end{figure}

\subsection{Ablation Study}
In this section, we explored the impact of various design choices, including prompting strategies, fine-tuning datasets, frame sampling strategies, and input resolution, on the performance of VLMs. 

\textbf{On prompting strategy.}
To isolate the contributions of each component in the SpatialMind prompting strategy, we evaluated two variants: one containing only the question component (CoT‑Question) and another containing only the scene description (CoT‑Scene). As shown in Table~\ref{tab:ablation}, both variants independently improve spatial reasoning performance,  but are less effective than the full combined prompt. Notably, the scene description contributes more significantly to model performance than the reasoning steps, suggesting its central role in facilitating spatial understanding.

\textbf{On fine-tuning data.}
To investigate the effectiveness of our proposed ScanForgeQA against existing spatial datasets, we fine-tuned Qwen2.5-VL-7B on SQA3D~\cite{masqa3d} and ScanQA~\cite{azuma2022scanqa}. As shown in Table~\ref{tab:ablation},  
fine-tuning on either of these datasets results in lower performance compared to ScanForgeQA, and even reduces accuracy on tasks involving precise spatial estimation (e.g., \textit{Room Size}).
This is primarily due to the limited presence of fine-grained spatial estimation samples in the existing datasets. 
Importantly, both datasets and the VSI-Bench benchmark originate from the same source (i.e., ScanNet~\cite{dai2017scannet}), resulting in minimal data discrepancy. This contrast emphasizes the advantage of our simulated data generation pipeline. 

\textbf{On frames and resolution.} To evaluate the robustness of our approach, we analyzed performance sensitivity to the number of input frames and image resolution.  Figure~\ref{fig:ablation} visualizes the performance of the +Both variant and the baseline  Qwen2.5-VL-7B under various configurations. Our method consistently outperforms the baseline across all settings, with performance further improving as the number of frames and resolution increase. 
This indicates that our approach remains stable and effective under varying visual input conditions.
\subsection{Qualitative Analysis}

\begin{figure}
  \centering
  \includegraphics[width=\linewidth]{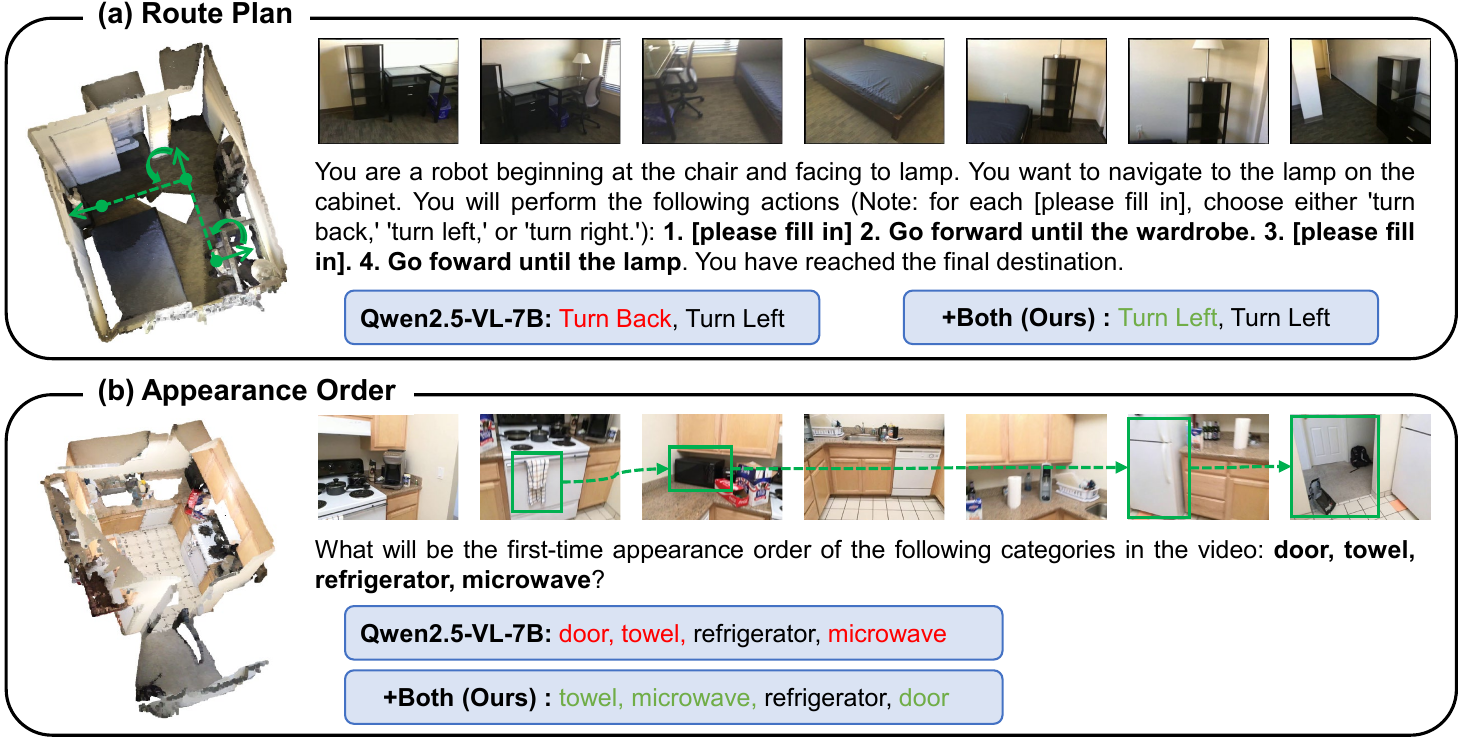}
  \caption{Two examples from VSI-Bench comparing predictions from Qwen2.5-VL-7B and Ours.}
  %\caption{Two cases from VSI-Bench.}
  \label{fig:case}
  \vspace{-2ex}
\end{figure}
In Figure~\ref{fig:case}, we presented two illustrative examples from VSI-Bench, comparing predictions from the baseline Qwen2.5-VL-7B and our enhanced variant (+Both). In Case (a), Qwen2.5-VL-7B fails to produce the correct directional prediction, likely due to its limited capacity for 3D spatial reasoning. In contrast, our method successfully identifies the correct answer. Case (b) involves a simpler spatial reasoning task,  however, Qwen2.5-VL-7B still fails, potentially due to insufficient object localization.  Our enhanced variant, benefiting from both structured prompting and spatially grounded fine-tuning, demonstrates notable improvements in accuracy and reasoning robustness.

% \vspace{1em} % 图与表之间的空隙

% \includegraphics[width=\linewidth]{img/gain.pdf} % 图像路径
% \captionof{table}{Ablation study on Qwen2.5-VL-7B with different frames and resolution.}\label{fig:gain}

\section{Conclusion}
In this work, we present an effective framework for enhancing visual-spatial reasoning in VLMs without modifying their underlying architecture. This makes our approach readily adaptable across models of varying scales and types. By integrating the structured prompting strategy (SpatialMind) with an automatically constructed dataset (ScanForgeQA), we enable VLMs to more effectively interpret and reason about 3D spatial relationships in complex visual scenes.  Extensive evaluations across multiple spatial reasoning benchmarks demonstrate that our framework consistently improves accuracy, robustness, and generalization. Furthermore, our analysis reveals that prompting and fine-tuning play complementary roles in advancing visual-spatial understanding.

% \begin{table}[ht]
% \centering
% \caption{Performance comparison on the EM-EQA subset of OpenEQA dataset.}
% \begin{tabular}{cccc}
% \toprule
% \textbf{Method} & \textbf{ScanNet} & \textbf{HM3D} & \textbf{All} \\
% \midrule
% Human Level & 87.7 & 85.1 & 86.8 \\
% GPT-4 & 32.5 & 35.5& 33.5\\
% GPT-4V & 57.4 & 51.3 & 55.3\\
% GPT-4o \\
%      +SpatialMind&\\
% Qwen2.5-VL-7B\\
%      +SpatialMind&\\
%      +ScanForgeQA&\\
%      \rowcolor{gray!20}+Both&\\
% \bottomrule
% \end{tabular}
% \label{tab:openeqa}
% \end{table}

% \begin{table}[ht]
% \centering
% \caption{Effects of different CoT methods.}
% \begin{tabular}{cc|cc}
% \toprule
% \textbf{Method} & \textbf{All} & \textbf{Method} & \textbf{All} \\
% \midrule
% Qwen2.5-VL-7B\\
% +CoT1&\\  
% +CoT2&\\  
% +SpatialMind&\\  
% InternVL2-7B\\
% +CoT1&&+Data1\\  
% +CoT2&&+Data2\\  
% +SpatialMind&\\  
% \bottomrule
% \end{tabular}
% \label{tab:cot}
% \end{table}

% \begin{ack}
% Use unnumbered first level headings for the acknowledgments. All acknowledgments

% Do {\bf not} include this section in the anonymized submission, only in the final paper. You can use the \texttt{ack} environment provided in the style file to automatically hide this section in the anonymized submission.
% \end{ack}

\bibliographystyle{unsrt}
\bibliography{reference}

\begin{thebibliography}{10}

\bibitem{yang2024thinking}
Jihan Yang, Shusheng Yang, Anjali~W Gupta, Rilyn Han, Li~Fei-Fei, and Saining Xie.
\newblock Thinking in space: How multimodal large language models see, remember, and recall spaces.
\newblock {\em arXiv preprint arXiv:2412.14171}, 2024.

\bibitem{zhang2023uncovering}
Haoyu Zhang, Meng Liu, Yaowei Wang, Da~Cao, Weili Guan, and Liqiang Nie.
\newblock Uncovering hidden connections: Iterative tracking and reasoning for video-grounded dialog.
\newblock {\em arXiv preprint arXiv:2310.07259}, 2023.

\bibitem{jin2024llava}
Yizhang Jin, Jian Li, Jiangning Zhang, Jianlong Hu, Zhenye Gan, Xin Tan, Yong Liu, Yabiao Wang, Chengjie Wang, and Lizhuang Ma.
\newblock Llava-vsd: Large language-and-vision assistant for visual spatial description.
\newblock In {\em Proceedings of the 32nd ACM International Conference on Multimedia}, pages 11420--11425, 2024.

\bibitem{zhang2023attribute}
Haoyu Zhang, Meng Liu, Yuhong Li, Ming Yan, Zan Gao, Xiaojun Chang, and Liqiang Nie.
\newblock Attribute-guided collaborative learning for partial person re-identification.
\newblock {\em IEEE Transactions on Pattern Analysis and Machine Intelligence}, 45(12):14144--14160, 2023.

\bibitem{tiandrivevlm}
Xiaoyu Tian, Junru Gu, Bailin Li, Yicheng Liu, Yang Wang, Zhiyong Zhao, Kun Zhan, Peng Jia, XianPeng Lang, and Hang Zhao.
\newblock Drivevlm: The convergence of autonomous driving and large vision-language models.
\newblock In {\em 8th Annual Conference on Robot Learning}.

\bibitem{zhang2021multimodal}
Haoyu Zhang, Meng Liu, Zan Gao, Xiaoqiang Lei, Yinglong Wang, and Liqiang Nie.
\newblock Multimodal dialog system: Relational graph-based context-aware question understanding.
\newblock In {\em Proceedings of the 29th ACM international conference on multimedia}, pages 695--703, 2021.

\bibitem{driess2023palm}
Danny Driess, Fei Xia, Mehdi~SM Sajjadi, Corey Lynch, Aakanksha Chowdhery, Brian Ichter, Ayzaan Wahid, Jonathan Tompson, Quan Vuong, Tianhe Yu, et~al.
\newblock Palm-e: an embodied multimodal language model.
\newblock In {\em Proceedings of the 40th International Conference on Machine Learning}, pages 8469--8488, 2023.

\bibitem{chandrasegaran2024hourvideo}
Keshigeyan Chandrasegaran, Agrim Gupta, Lea~M Hadzic, Taran Kota, Jimming He, Crist{\'o}bal Eyzaguirre, Zane Durante, Manling Li, Jiajun Wu, and Fei-Fei Li.
\newblock Hourvideo: 1-hour video-language understanding.
\newblock {\em Advances in Neural Information Processing Systems}, 37:53168--53197, 2024.

\bibitem{zhang2025exo2ego}
Haoyu Zhang, Qiaohui Chu, Meng Liu, Yunxiao Wang, Bin Wen, Fan Yang, Tingting Gao, Di~Zhang, Yaowei Wang, and Liqiang Nie.
\newblock Exo2ego: Exocentric knowledge guided mllm for egocentric video understanding.
\newblock {\em arXiv preprint arXiv:2503.09143}, 2025.

\bibitem{feng2024objectnlq}
Yisen Feng, Haoyu Zhang, Yuquan Xie, Zaijing Li, Meng Liu, and Liqiang Nie.
\newblock Objectnlq@ ego4d episodic memory challenge 2024.
\newblock {\em arXiv preprint arXiv:2406.15778}, 2024.

\bibitem{feng2025object}
Yisen Feng, Haoyu Zhang, Meng Liu, Weili Guan, and Liqiang Nie.
\newblock Object-shot enhanced grounding network for egocentric video.
\newblock {\em arXiv preprint arXiv:2505.04270}, 2025.

\bibitem{chu2025technicalreportego4dlongterm}
Qiaohui Chu, Haoyu Zhang, Yisen Feng, Meng Liu, Weili Guan, Yaowei Wang, and Liqiang Nie.
\newblock Technical report for ego4d long-term action anticipation challenge 2025.
\newblock {\em arXiv preprint arXiv:2506.02550}, 2025.

\bibitem{chen2024spatialvlm}
Boyuan Chen, Zhuo Xu, Sean Kirmani, Brain Ichter, Dorsa Sadigh, Leonidas Guibas, and Fei Xia.
\newblock Spatialvlm: Endowing vision-language models with spatial reasoning capabilities.
\newblock In {\em Proceedings of the IEEE/CVF Conference on Computer Vision and Pattern Recognition}, pages 14455--14465, 2024.

\bibitem{cheng2025spatialrgpt}
An-Chieh Cheng, Hongxu Yin, Yang Fu, Qiushan Guo, Ruihan Yang, Jan Kautz, Xiaolong Wang, and Sifei Liu.
\newblock Spatialrgpt: Grounded spatial reasoning in vision-language models.
\newblock {\em Advances in Neural Information Processing Systems}, 37:135062--135093, 2025.

\bibitem{cai2024spatialbot}
Wenxiao Cai, Iaroslav Ponomarenko, Jianhao Yuan, Xiaoqi Li, Wankou Yang, Hao Dong, and Bo~Zhao.
\newblock Spatialbot: Precise spatial understanding with vision language models.
\newblock {\em arXiv preprint arXiv:2406.13642}, 2024.

\bibitem{chen2024ll3da}
Sijin Chen, Xin Chen, Chi Zhang, Mingsheng Li, Gang Yu, Hao Fei, Hongyuan Zhu, Jiayuan Fan, and Tao Chen.
\newblock Ll3da: Visual interactive instruction tuning for omni-3d understanding reasoning and planning.
\newblock In {\em Proceedings of the IEEE/CVF Conference on Computer Vision and Pattern Recognition}, pages 26428--26438, 2024.

\bibitem{hong20233d}
Yining Hong, Haoyu Zhen, Peihao Chen, Shuhong Zheng, Yilun Du, Zhenfang Chen, and Chuang Gan.
\newblock 3d-llm: Injecting the 3d world into large language models.
\newblock {\em Advances in Neural Information Processing Systems}, 36:20482--20494, 2023.

\bibitem{qi2025gpt4scene}
Zhangyang Qi, Zhixiong Zhang, Ye~Fang, Jiaqi Wang, and Hengshuang Zhao.
\newblock Gpt4scene: Understand 3d scenes from videos with vision-language models.
\newblock {\em arXiv preprint arXiv:2501.01428}, 2025.

\bibitem{pmlr-v235-zhang24aj}
Haoyu Zhang, Meng Liu, Zixin Liu, Xuemeng Song, Yaowei Wang, and Liqiang Nie.
\newblock Multi-factor adaptive vision selection for egocentric video question answering.
\newblock In {\em Proceedings of the 41st International Conference on Machine Learning}, volume 235 of {\em Proceedings of Machine Learning Research}, pages 59310--59328. PMLR, 2024.

\bibitem{du2024embspatial}
Mengfei Du, Binhao Wu, Zejun Li, Xuan-Jing Huang, and Zhongyu Wei.
\newblock Embspatial-bench: Benchmarking spatial understanding for embodied tasks with large vision-language models.
\newblock In {\em Proceedings of the 62nd Annual Meeting of the Association for Computational Linguistics (Volume 2: Short Papers)}, pages 346--355, 2024.

\bibitem{zhang2024sphere}
Wenyu Zhang, Wei~En Ng, Lixin Ma, Yuwen Wang, Jungqi Zhao, Allison Koenecke, Boyang Li, and Lu~Wang.
\newblock Sphere: Unveiling spatial blind spots in vision-language models through hierarchical evaluation.
\newblock {\em arXiv preprint arXiv:2412.12693}, 2024.

\bibitem{maspatialpin}
Chenyang Ma, Kai Lu, Ta-Ying Cheng, Niki Trigoni, and Andrew Markham.
\newblock Spatialpin: Enhancing spatial reasoning capabilities of vision-language models through prompting and interacting 3d priors.
\newblock In {\em The Thirty-eighth Annual Conference on Neural Information Processing Systems}.

\bibitem{yu2024rag}
Jun Yu, Yunxiang Zhang, Zerui Zhang, Zhao Yang, Gongpeng Zhao, Fengzhao Sun, Fanrui Zhang, Qingsong Liu, Jianqing Sun, Jiaen Liang, et~al.
\newblock Rag-guided large language models for visual spatial description with adaptive hallucination corrector.
\newblock In {\em Proceedings of the 32nd ACM International Conference on Multimedia}, pages 11407--11413, 2024.

\bibitem{yuanrobopoint}
Wentao Yuan, Jiafei Duan, Valts Blukis, Wilbert Pumacay, Ranjay Krishna, Adithyavairavan Murali, Arsalan Mousavian, and Dieter Fox.
\newblock Robopoint: A vision-language model for spatial affordance prediction in robotics.
\newblock In {\em 8th Annual Conference on Robot Learning}.

\bibitem{song2024robospatial}
Chan~Hee Song, Valts Blukis, Jonathan Tremblay, Stephen Tyree, Yu~Su, and Stan Birchfield.
\newblock Robospatial: Teaching spatial understanding to 2d and 3d vision-language models for robotics.
\newblock {\em arXiv preprint arXiv:2411.16537}, 2024.

\bibitem{liu2025spatialcot}
Yuecheng Liu, Dafeng Chi, Shiguang Wu, Zhanguang Zhang, Yaochen Hu, Lingfeng Zhang, Yingxue Zhang, Shuang Wu, Tongtong Cao, Guowei Huang, et~al.
\newblock Spatialcot: Advancing spatial reasoning through coordinate alignment and chain-of-thought for embodied task planning.
\newblock {\em arXiv preprint arXiv:2501.10074}, 2025.

\bibitem{zhang2024hcqa}
Haoyu Zhang, Yuquan Xie, Yisen Feng, Zaijing Li, Meng Liu, and Liqiang Nie.
\newblock Hcqa@ ego4d egoschema challenge 2024.
\newblock {\em arXiv preprint arXiv:2406.15771}, 2024.

\bibitem{zhang2025hcqa}
Haoyu Zhang, Yisen Feng, Qiaohui Chu, Meng Liu, Weili Guan, Yaowei Wang, and Liqiang Nie.
\newblock Hcqa-1.5@ ego4d egoschema challenge 2025.
\newblock {\em arXiv preprint arXiv:2505.20644}, 2025.

\bibitem{baruch1arkitscenes}
Gilad Baruch, Zhuoyuan Chen, Afshin Dehghan, Yuri Feigin, Peter Fu, Thomas Gebauer, Daniel Kurz, Tal Dimry, Brandon Joffe, Arik Schwartz, et~al.
\newblock Arkitscenes: A diverse real-world dataset for 3d indoor scene understanding using mobile rgb-d data.
\newblock In {\em Thirty-fifth Conference on Neural Information Processing Systems Datasets and Benchmarks Track (Round 1)}.

\bibitem{chang2017matterport3d}
Angel Chang, Angela Dai, Thomas Funkhouser, Maciej Halber, Matthias Niebner, Manolis Savva, Shuran Song, Andy Zeng, and Yinda Zhang.
\newblock Matterport3d: Learning from rgb-d data in indoor environments.
\newblock In {\em International Conference on 3D Vision (3DV)}, 2017.

\bibitem{dai2017scannet}
Angela Dai, Angel~X Chang, Manolis Savva, Maciej Halber, Thomas Funkhouser, and Matthias Nie{\ss}ner.
\newblock Scannet: Richly-annotated 3d reconstructions of indoor scenes.
\newblock In {\em Proceedings of the IEEE conference on computer vision and pattern recognition}, pages 5828--5839, 2017.

\bibitem{deitke2022️}
Matt Deitke, Eli VanderBilt, Alvaro Herrasti, Luca Weihs, Kiana Ehsani, Jordi Salvador, Winson Han, Eric Kolve, Aniruddha Kembhavi, and Roozbeh Mottaghi.
\newblock Procthor: Large-scale embodied ai using procedural generation.
\newblock {\em Advances in Neural Information Processing Systems}, 35:5982--5994, 2022.

\bibitem{mao2022multiscan}
Yongsen Mao, Yiming Zhang, Hanxiao Jiang, Angel Chang, and Manolis Savva.
\newblock Multiscan: Scalable rgbd scanning for 3d environments with articulated objects.
\newblock {\em Advances in neural information processing systems}, 35:9058--9071, 2022.

\bibitem{ramakrishnan2habitat}
Santhosh~Kumar Ramakrishnan, Aaron Gokaslan, Erik Wijmans, Oleksandr Maksymets, Alexander Clegg, John~M Turner, Eric Undersander, Wojciech Galuba, Andrew Westbury, Angel~X Chang, et~al.
\newblock Habitat-matterport 3d dataset (hm3d): 1000 large-scale 3d environments for embodied ai.
\newblock In {\em Thirty-fifth Conference on Neural Information Processing Systems Datasets and Benchmarks Track (Round 2)}.

\bibitem{yeshwanth2023scannet++}
Chandan Yeshwanth, Yueh-Cheng Liu, Matthias Nie{\ss}ner, and Angela Dai.
\newblock Scannet++: A high-fidelity dataset of 3d indoor scenes.
\newblock In {\em Proceedings of the IEEE/CVF International Conference on Computer Vision}, pages 12--22, 2023.

\bibitem{vu2022softgroup}
Thang Vu, Kookhoi Kim, Tung~M Luu, Thanh Nguyen, and Chang~D Yoo.
\newblock Softgroup for 3d instance segmentation on point clouds.
\newblock In {\em Proceedings of the IEEE/CVF conference on computer vision and pattern recognition}, pages 2708--2717, 2022.

\bibitem{wu2024point}
Xiaoyang Wu, Li~Jiang, Peng-Shuai Wang, Zhijian Liu, Xihui Liu, Yu~Qiao, Wanli Ouyang, Tong He, and Hengshuang Zhao.
\newblock Point transformer v3: Simpler faster stronger.
\newblock In {\em Proceedings of the IEEE/CVF Conference on Computer Vision and Pattern Recognition}, pages 4840--4851, 2024.

\bibitem{nguyen2024open3dis}
Phuc Nguyen, Tuan~Duc Ngo, Evangelos Kalogerakis, Chuang Gan, Anh Tran, Cuong Pham, and Khoi Nguyen.
\newblock Open3dis: Open-vocabulary 3d instance segmentation with 2d mask guidance.
\newblock In {\em Proceedings of the IEEE/CVF Conference on Computer Vision and Pattern Recognition}, pages 4018--4028, 2024.

\bibitem{rozenberszki2024unscene3d}
David Rozenberszki, Or~Litany, and Angela Dai.
\newblock Unscene3d: Unsupervised 3d instance segmentation for indoor scenes.
\newblock In {\em Proceedings of the IEEE/CVF Conference on Computer Vision and Pattern Recognition}, pages 19957--19967, 2024.

\bibitem{guo2023point}
Ziyu Guo, Renrui Zhang, Xiangyang Zhu, Yiwen Tang, Xianzheng Ma, Jiaming Han, Kexin Chen, Peng Gao, Xianzhi Li, Hongsheng Li, et~al.
\newblock Point-bind \& point-llm: Aligning point cloud with multi-modality for 3d understanding, generation, and instruction following.
\newblock {\em arXiv preprint arXiv:2309.00615}, 2023.

\bibitem{qi2024shapellm}
Zekun Qi, Runpei Dong, Shaochen Zhang, Haoran Geng, Chunrui Han, Zheng Ge, Li~Yi, and Kaisheng Ma.
\newblock Shapellm: Universal 3d object understanding for embodied interaction.
\newblock In {\em European Conference on Computer Vision}, pages 214--238. Springer, 2024.

\bibitem{liu2024lion}
Zhe Liu, Jinghua Hou, Xinyu Wang, Xiaoqing Ye, Jingdong Wang, Hengshuang Zhao, and Xiang Bai.
\newblock Lion: Linear group rnn for 3d object detection in point clouds.
\newblock {\em Advances in Neural Information Processing Systems}, 37:13601--13626, 2024.

\bibitem{xu2024pointllm}
Runsen Xu, Xiaolong Wang, Tai Wang, Yilun Chen, Jiangmiao Pang, and Dahua Lin.
\newblock Pointllm: Empowering large language models to understand point clouds.
\newblock In {\em European Conference on Computer Vision}, pages 131--147. Springer, 2024.

\bibitem{man2024lexicon3d}
Yunze Man, Shuhong Zheng, Zhipeng Bao, Martial Hebert, Liangyan Gui, and Yu-Xiong Wang.
\newblock Lexicon3d: Probing visual foundation models for complex 3d scene understanding.
\newblock {\em Advances in Neural Information Processing Systems}, 37:76819--76847, 2024.

\bibitem{liao2025improved}
Zhenyi Liao, Qingsong Xie, Yanhao Zhang, Zijian Kong, Haonan Lu, Zhenyu Yang, and Zhijie Deng.
\newblock Improved visual-spatial reasoning via r1-zero-like training.
\newblock {\em arXiv preprint arXiv:2504.00883}, 2025.

\bibitem{fu20213d}
Huan Fu, Bowen Cai, Lin Gao, Ling-Xiao Zhang, Jiaming Wang, Cao Li, Qixun Zeng, Chengyue Sun, Rongfei Jia, Binqiang Zhao, et~al.
\newblock 3d-front: 3d furnished rooms with layouts and semantics.
\newblock In {\em Proceedings of the IEEE/CVF International Conference on Computer Vision}, pages 10933--10942, 2021.

\bibitem{yang2024holodeck}
Yue Yang, Fan-Yun Sun, Luca Weihs, Eli VanderBilt, Alvaro Herrasti, Winson Han, Jiajun Wu, Nick Haber, Ranjay Krishna, Lingjie Liu, et~al.
\newblock Holodeck: Language guided generation of 3d embodied ai environments.
\newblock In {\em Proceedings of the IEEE/CVF Conference on Computer Vision and Pattern Recognition}, pages 16227--16237, 2024.

\bibitem{deitke2023objaverse}
Matt Deitke, Dustin Schwenk, Jordi Salvador, Luca Weihs, Oscar Michel, Eli VanderBilt, Ludwig Schmidt, Kiana Ehsani, Aniruddha Kembhavi, and Ali Farhadi.
\newblock Objaverse: A universe of annotated 3d objects.
\newblock In {\em Proceedings of the IEEE/CVF conference on computer vision and pattern recognition}, pages 13142--13153, 2023.

\bibitem{majumdar2024openeqa}
Arjun Majumdar, Anurag Ajay, Xiaohan Zhang, Pranav Putta, Sriram Yenamandra, Mikael Henaff, Sneha Silwal, Paul Mcvay, Oleksandr Maksymets, Sergio Arnaud, et~al.
\newblock Openeqa: Embodied question answering in the era of foundation models.
\newblock In {\em Proceedings of the IEEE/CVF conference on computer vision and pattern recognition}, pages 16488--16498, 2024.

\bibitem{azuma2022scanqa}
Daichi Azuma, Taiki Miyanishi, Shuhei Kurita, and Motoaki Kawanabe.
\newblock Scanqa: 3d question answering for spatial scene understanding.
\newblock In {\em proceedings of the IEEE/CVF conference on computer vision and pattern recognition}, pages 19129--19139, 2022.

\bibitem{masqa3d}
Xiaojian Ma, Silong Yong, Zilong Zheng, Qing Li, Yitao Liang, Song-Chun Zhu, and Siyuan Huang.
\newblock Sqa3d: Situated question answering in 3d scenes.
\newblock In {\em The Eleventh International Conference on Learning Representations}.

\bibitem{li2024mvbench}
Kunchang Li, Yali Wang, Yinan He, Yizhuo Li, Yi~Wang, Yi~Liu, Zun Wang, Jilan Xu, Guo Chen, Ping Luo, et~al.
\newblock Mvbench: A comprehensive multi-modal video understanding benchmark.
\newblock In {\em Proceedings of the IEEE/CVF Conference on Computer Vision and Pattern Recognition}, pages 22195--22206, 2024.

\bibitem{fu2024video}
Chaoyou Fu, Yuhan Dai, Yongdong Luo, Lei Li, Shuhuai Ren, Renrui Zhang, Zihan Wang, Chenyu Zhou, Yunhang Shen, Mengdan Zhang, et~al.
\newblock Video-mme: The first-ever comprehensive evaluation benchmark of multi-modal llms in video analysis.
\newblock {\em arXiv preprint arXiv:2405.21075}, 2024.

\bibitem{chen2024sharegpt4video}
Lin Chen, Xilin Wei, Jinsong Li, Xiaoyi Dong, Pan Zhang, Yuhang Zang, Zehui Chen, Haodong Duan, Zhenyu Tang, Li~Yuan, et~al.
\newblock Sharegpt4video: Improving video understanding and generation with better captions.
\newblock {\em Advances in Neural Information Processing Systems}, 37:19472--19495, 2024.

\bibitem{chen2024expanding}
Zhe Chen, Weiyun Wang, Yue Cao, Yangzhou Liu, Zhangwei Gao, Erfei Cui, Jinguo Zhu, Shenglong Ye, Hao Tian, Zhaoyang Liu, et~al.
\newblock Expanding performance boundaries of open-source multimodal models with model, data, and test-time scaling.
\newblock {\em arXiv preprint arXiv:2412.05271}, 2024.

\bibitem{li2024llava}
Bo~Li, Yuanhan Zhang, Dong Guo, Renrui Zhang, Feng Li, Hao Zhang, Kaichen Zhang, Peiyuan Zhang, Yanwei Li, Ziwei Liu, et~al.
\newblock Llava-onevision: Easy visual task transfer.
\newblock {\em arXiv preprint arXiv:2408.03326}, 2024.

\bibitem{zhang2025videollama}
Boqiang Zhang, Kehan Li, Zesen Cheng, Zhiqiang Hu, Yuqian Yuan, Guanzheng Chen, Sicong Leng, Yuming Jiang, Hang Zhang, Xin Li, et~al.
\newblock Videollama 3: Frontier multimodal foundation models for image and video understanding.
\newblock {\em arXiv preprint arXiv:2501.13106}, 2025.

\bibitem{wang2024qwen2}
Peng Wang, Shuai Bai, Sinan Tan, Shijie Wang, Zhihao Fan, Jinze Bai, Keqin Chen, Xuejing Liu, Jialin Wang, Wenbin Ge, et~al.
\newblock Qwen2-vl: Enhancing vision-language model's perception of the world at any resolution.
\newblock {\em arXiv preprint arXiv:2409.12191}, 2024.

\bibitem{bai2025qwen2}
Shuai Bai, Keqin Chen, Xuejing Liu, Jialin Wang, Wenbin Ge, Sibo Song, Kai Dang, Peng Wang, Shijie Wang, Jun Tang, et~al.
\newblock Qwen2. 5-vl technical report.
\newblock {\em arXiv preprint arXiv:2502.13923}, 2025.

\bibitem{hurst2024gpt}
Aaron Hurst, Adam Lerer, Adam~P Goucher, Adam Perelman, Aditya Ramesh, Aidan Clark, AJ~Ostrow, Akila Welihinda, Alan Hayes, Alec Radford, et~al.
\newblock Gpt-4o system card.
\newblock {\em arXiv preprint arXiv:2410.21276}, 2024.

\bibitem{schuhmann2022laion}
Christoph Schuhmann, Romain Beaumont, Richard Vencu, Cade Gordon, Ross Wightman, Mehdi Cherti, Theo Coombes, Aarush Katta, Clayton Mullis, Mitchell Wortsman, et~al.
\newblock Laion-5b: An open large-scale dataset for training next generation image-text models.
\newblock {\em Advances in neural information processing systems}, 35:25278--25294, 2022.

\bibitem{yang2024vision}
Chenyu Yang, Xizhou Zhu, Jinguo Zhu, Weijie Su, Junjie Wang, Xuan Dong, Wenhai Wang, Bin Li, Jie Zhou, Yu~Qiao, et~al.
\newblock Vision model pre-training on interleaved image-text data via latent compression learning.
\newblock {\em Advances in Neural Information Processing Systems}, 37:23912--23938, 2024.

\bibitem{guan2022bi}
Weili Guan, Xuemeng Song, Haoyu Zhang, Meng Liu, Chung-Hsing Yeh, and Xiaojun Chang.
\newblock Bi-directional heterogeneous graph hashing towards efficient outfit recommendation.
\newblock In {\em Proceedings of the 30th ACM international conference on multimedia}, pages 268--276, 2022.

\bibitem{alayrac2022flamingo}
Jean-Baptiste Alayrac, Jeff Donahue, Pauline Luc, Antoine Miech, Iain Barr, Yana Hasson, Karel Lenc, Arthur Mensch, Katherine Millican, Malcolm Reynolds, et~al.
\newblock Flamingo: a visual language model for few-shot learning.
\newblock {\em Advances in neural information processing systems}, 35:23716--23736, 2022.

\bibitem{huang2023language}
Shaohan Huang, Li~Dong, Wenhui Wang, Yaru Hao, Saksham Singhal, Shuming Ma, Tengchao Lv, Lei Cui, Owais~Khan Mohammed, Barun Patra, et~al.
\newblock Language is not all you need: Aligning perception with language models.
\newblock {\em Advances in Neural Information Processing Systems}, 36:72096--72109, 2023.

\bibitem{li2022blip}
Junnan Li, Dongxu Li, Caiming Xiong, and Steven Hoi.
\newblock Blip: Bootstrapping language-image pre-training for unified vision-language understanding and generation.
\newblock In {\em International conference on machine learning}, pages 12888--12900. PMLR, 2022.

\bibitem{li2023blip}
Junnan Li, Dongxu Li, Silvio Savarese, and Steven Hoi.
\newblock Blip-2: Bootstrapping language-image pre-training with frozen image encoders and large language models.
\newblock In {\em International conference on machine learning}, pages 19730--19742. PMLR, 2023.

\bibitem{liu2023visual}
Haotian Liu, Chunyuan Li, Qingyang Wu, and Yong~Jae Lee.
\newblock Visual instruction tuning.
\newblock {\em Advances in neural information processing systems}, 36:34892--34916, 2023.

\bibitem{zhu2023minigpt}
Deyao Zhu, Jun Chen, Xiaoqian Shen, Xiang Li, and Mohamed Elhoseiny.
\newblock Minigpt-4: Enhancing vision-language understanding with advanced large language models.
\newblock {\em arXiv preprint arXiv:2304.10592}, 2023.

\bibitem{liu2024improved}
Haotian Liu, Chunyuan Li, Yuheng Li, and Yong~Jae Lee.
\newblock Improved baselines with visual instruction tuning.
\newblock In {\em Proceedings of the IEEE/CVF Conference on Computer Vision and Pattern Recognition}, pages 26296--26306, 2024.

\bibitem{chen2023minigpt}
Jun Chen, Deyao Zhu, Xiaoqian Shen, Xiang Li, Zechun Liu, Pengchuan Zhang, Raghuraman Krishnamoorthi, Vikas Chandra, Yunyang Xiong, and Mohamed Elhoseiny.
\newblock Minigpt-v2: large language model as a unified interface for vision-language multi-task learning.
\newblock {\em arXiv preprint arXiv:2310.09478}, 2023.

\bibitem{lai2024lisa}
Xin Lai, Zhuotao Tian, Yukang Chen, Yanwei Li, Yuhui Yuan, Shu Liu, and Jiaya Jia.
\newblock Lisa: Reasoning segmentation via large language model.
\newblock In {\em Proceedings of the IEEE/CVF Conference on Computer Vision and Pattern Recognition}, pages 9579--9589, 2024.

\bibitem{sunemu}
Quan Sun, Qiying Yu, Yufeng Cui, Fan Zhang, Xiaosong Zhang, Yueze Wang, Hongcheng Gao, Jingjing Liu, Tiejun Huang, and Xinlong Wang.
\newblock Emu: Generative pretraining in multimodality.
\newblock In {\em The Twelfth International Conference on Learning Representations}, 2024.

\bibitem{wang2024cogvlm}
Weihan Wang, Qingsong Lv, Wenmeng Yu, Wenyi Hong, Ji~Qi, Yan Wang, Junhui Ji, Zhuoyi Yang, Lei Zhao, Song XiXuan, et~al.
\newblock Cogvlm: Visual expert for pretrained language models.
\newblock {\em Advances in Neural Information Processing Systems}, 37:121475--121499, 2024.

\bibitem{xiao2024can}
Junbin Xiao, Angela Yao, Yicong Li, and Tat-Seng Chua.
\newblock Can i trust your answer? visually grounded video question answering.
\newblock In {\em Proceedings of the IEEE/CVF Conference on Computer Vision and Pattern Recognition}, pages 13204--13214, 2024.

\bibitem{liu2018attentive}
Meng Liu, Xiang Wang, Liqiang Nie, Xiangnan He, Baoquan Chen, and Tat-Seng Chua.
\newblock Attentive moment retrieval in videos.
\newblock In {\em The 41st international ACM SIGIR conference on research \& development in information retrieval}, pages 15--24, 2018.

\bibitem{liu2018cross}
Meng Liu, Xiang Wang, Liqiang Nie, Qi~Tian, Baoquan Chen, and Tat-Seng Chua.
\newblock Cross-modal moment localization in videos.
\newblock In {\em Proceedings of the 26th ACM international conference on Multimedia}, pages 843--851, 2018.

\bibitem{cheng2024videollama}
Zesen Cheng, Sicong Leng, Hang Zhang, Yifei Xin, Xin Li, Guanzheng Chen, Yongxin Zhu, Wenqi Zhang, Ziyang Luo, Deli Zhao, et~al.
\newblock Videollama 2: Advancing spatial-temporal modeling and audio understanding in video-llms.
\newblock {\em arXiv preprint arXiv:2406.07476}, 2024.

\bibitem{qian2024streaming}
Rui Qian, Xiaoyi Dong, Pan Zhang, Yuhang Zang, Shuangrui Ding, Dahua Lin, and Jiaqi Wang.
\newblock Streaming long video understanding with large language models.
\newblock {\em Advances in Neural Information Processing Systems}, 37:119336--119360, 2024.

\bibitem{xing2024pyramiddrop}
Long Xing, Qidong Huang, Xiaoyi Dong, Jiajie Lu, Pan Zhang, Yuhang Zang, Yuhang Cao, Conghui He, Jiaqi Wang, Feng Wu, et~al.
\newblock Pyramiddrop: Accelerating your large vision-language models via pyramid visual redundancy reduction.
\newblock In {\em Proceedings of the IEEE/CVF Conference on Computer Vision and Pattern Recognition}, pages 1--13, 2025.

\bibitem{zhang2024sparsevlm}
Yuan Zhang, Chun-Kai Fan, Junpeng Ma, Wenzhao Zheng, Tao Huang, Kuan Cheng, Denis Gudovskiy, Tomoyuki Okuno, Yohei Nakata, Kurt Keutzer, et~al.
\newblock Sparsevlm: Visual token sparsification for efficient vision-language model inference.
\newblock {\em arXiv preprint arXiv:2410.04417}, 2024.

\bibitem{wan2024look}
Zhongwei Wan, Ziang Wu, Che Liu, Jinfa Huang, Zhihong Zhu, Peng Jin, Longyue Wang, and Li~Yuan.
\newblock Look-m: Look-once optimization in kv cache for efficient multimodal long-context inference.
\newblock In {\em Findings of the Association for Computational Linguistics: EMNLP 2024}, pages 4065--4078, 2024.

\bibitem{wang2025time}
Yunxiao Wang, Meng Liu, Rui Shao, Haoyu Zhang, Bin Wen, Fan Yang, Tingting Gao, Di~Zhang, and Liqiang Nie.
\newblock Time: Temporal-sensitive multi-dimensional instruction tuning and benchmarking for video-llms.
\newblock {\em arXiv preprint arXiv:2503.09994}, 2025.

\bibitem{li2025sti}
Yun Li, Yiming Zhang, Tao Lin, XiangRui Liu, Wenxiao Cai, Zheng Liu, and Bo~Zhao.
\newblock Sti-bench: Are mllms ready for precise spatial-temporal world understanding?
\newblock {\em arXiv preprint arXiv:2503.23765}, 2025.

\bibitem{zhang2024agent3d}
Sha Zhang, Di~Huang, Jiajun Deng, Shixiang Tang, Wanli Ouyang, Tong He, and Yanyong Zhang.
\newblock Agent3d-zero: An agent for zero-shot 3d understanding.
\newblock In {\em European Conference on Computer Vision}, pages 186--202. Springer, 2024.

\bibitem{zheng2024video}
Duo Zheng, Shijia Huang, and Liwei Wang.
\newblock Video-3d llm: Learning position-aware video representation for 3d scene understanding.
\newblock In {\em Proceedings of the IEEE/CVF Conference on Computer Vision and Pattern Recognition}, pages 1--14, 2025.

\end{thebibliography}

% \section*{References}

% References follow the acknowledgments in the camera-ready paper. Use unnumbered first-level heading for
% the references. Any choice of citation style is acceptable as long as you are
% consistent. It is permissible to reduce the font size to \verb+small+ (9 point)
% when listing the references.
% Note that the Reference section does not count towards the page limit.
% \medskip

% {
% \small

% [1] Alexander, J.A.\ \& Mozer, M.C.\ (1995) Template-based algorithms for
% connectionist rule extraction. In G.\ Tesauro, D.S.\ Touretzky and T.K.\ Leen
% (eds.), {\it Advances in Neural Information Processing Systems 7},
% pp.\ 609--616. Cambridge, MA: MIT Press.

% [2] Bower, J.M.\ \& Beeman, D.\ (1995) {\it The Book of GENESIS: Exploring
%   Realistic Neural Models with the GEneral NEural SImulation System.}  New York:
% TELOS/Springer--Verlag.

% [3] Hasselmo, M.E., Schnell, E.\ \& Barkai, E.\ (1995) Dynamics of learning and
% recall at excitatory recurrent synapses and cholinergic modulation in rat
% hippocampal region CA3. {\it Journal of Neuroscience} {\bf 15}(7):5249-5262.
% }

%%%%%%%%%%%%%%%%%%%%%%%%%%%%%%%%%%%%%%%%%%%%%%%%%%%%%%%%%%%%
\newpage
\renewcommand{\cftsecfont}{\color{blue}} % 一级标题颜色
\renewcommand{\cftsubsecfont}{\color{blue}} % 二级标题颜色
\renewcommand{\cftsubsubsecfont}{\color{blue}} % 二级标题颜色
\renewcommand{\cftsecpagefont}{\color{black}} % 页码颜色
\renewcommand{\cftsubsecpagefont}{\color{black}} % 页码颜色

\begin{center}
    \LARGE \textbf{Appendix}
\end{center}

\tableofcontents

\appendix

\newpage

\section{Experimental Settings}\label{app_setting}

\subsection{Benchmarks}
To validate the effectiveness of our SpatialMind strategy and ScanForgeQA data in enhancing the visual-spatial understanding capabilities of VLMs,
we conducted evaluations on four comprehensive benchmarks: VSI-Bench~\cite{yang2024thinking}, OpenEQA~\cite{majumdar2024openeqa}, ScanQA~\cite{azuma2022scanqa}, and SQA3D~\cite{masqa3d}. The comparisons are shown in Table~\ref{1}.

\begin{table}[h]
\centering
\caption{Comparison of different benchmarks. Note that the statistics for the ScanQA and SQA3D datasets are reported on their respective validation sets.}
  \renewcommand{\arraystretch}{1.1} % 默认是1.0，1.5表示增加50%行高
\begin{tabular}{M{1.5cm}M{1.5cm}M{1.5cm}M{2cm}M{5.5cm}}
\hline
\textbf{Dataset} & \textbf{\#Questions} & \textbf{\#Scenes} & \textbf{Source} & \textbf{Question Types} \\
\hline
VSI-Bench & $>$5,000 & 288 & ScanNet, ScanNet++, ARKitScenes & Configuration, Measurement Estimation, Spatio-temporal Reasoning \\
\hline
OpenEQA  & 1,899 & $>$180 & ScanNet, HM3D & Object Recognition, Attribute Reasoning, Spatial Understanding, Functional Reasoning \\
\hline
ScanQA    & 9,353   & 71 & ScanNet & Object, Color, Quantity, Location \\
\hline
SQA3D     & 3,261   & 65 & ScanNet & Spatial Relations, Commonsense, Navigation, Multi-hop Reasoning \\
\hline
\end{tabular}\label{1}

\end{table}

\subsubsection{VSI-Bench}
% consists of over 5,000 QA pairs grounded in 288 real-world egocentric videos, sourced from the validation sets of ScanNet~\cite{dai2017scannet}, ScanNet++~\cite{yeshwanth2023scannet++}, and ARKitScenes~\cite{baruch1arkitscenes}. These videos span diverse environments—including residential, professional, and industrial settings—and offer high-quality object-level annotations inherited from 3D reconstruction datasets. The benchmark has undergone iterative refinement to ensure question clarity and annotation accuracy. 
VSI-Bench~\cite{yang2024thinking} is a comprehensive evaluation benchmark designed to assess the visual-spatial reasoning capabilities of VLMs in dynamic 3D environments. It comprises over 5,000 high-quality question–answer pairs grounded in 288 diverse indoor video scenes, sourced from real-world 3D reconstruction datasets such as ScanNet~\cite{dai2017scannet}, ScanNet++~\cite{yeshwanth2023scannet++}, and ARKitScenes~\cite{baruch1arkitscenes}. The benchmark defines eight distinct tasks organized into three major categories: \textbf{configuration-based tasks} (e.g., object counting, relative direction, and path planning), \textbf{measurement estimation tasks} (e.g., object size, absolute distance, room size), and \textbf{spatiotemporal reasoning tasks} (e.g., object appearance order). These tasks collectively assess VLMs' abilities to reason about spatial relationships, estimate scale, and understand temporal information. Evaluation is primarily based on accuracy, with tolerance-based error metrics applied to numerical estimation tasks to account for minor deviations. All evaluations follow the official  protocols defined in VSI-Bench~\cite{yang2024thinking}.
\begin{table}
  \caption{Performance comparison on VSI-Bench. \textcolor{red}{\raisebox{0.2ex}{$\dagger$}} indicates results on VSI-Bench (tiny) set and LLaVA-NV denotes the LLaVA-NeXT-Video model.}
  \renewcommand{\arraystretch}{1.1} % 默认是1.0，1.5表示增加50%行高

  \centering
  \setlength{\tabcolsep}{4pt} % 缩小列间距
  \begin{tabular}{lcccccccccc}
    \toprule
    \textbf{Method} & 
    \makecell{\textbf{Obj.} \\ \textbf{Count}} & 
    \makecell{\textbf{Abs.} \\ \textbf{Dist.}} & 
    \makecell{\textbf{Obj.} \\ \textbf{Size}} & 
    \makecell{\textbf{Room} \\ \textbf{Size}} & 
    \makecell{\textbf{Rel.} \\ \textbf{Dist.}} & 
    \makecell{\textbf{Rel.} \\ \textbf{Dir.}} & 
    \makecell{\textbf{Route} \\ \textbf{Plan}} & 
    \makecell{\textbf{Appr.} \\ \textbf{Order}} & 
    \textbf{Avg}&
    $\Delta$\\

    % \textbf{Method} & \textbf{Obj. Count} & \textbf{Abs. Dist.} & \textbf{Obj. Size} & \textbf{Room Size} & \textbf{Rel. Dist.} & \textbf{Rel. Dir.}$^{\star}$ & \textbf{Room Plan} & \textbf{Appr. Order} & \textbf{Avg.}\\
    \midrule
    \multicolumn{11}{c}{Close-source}\\
    \midrule
     Human Level\textcolor{red}{\raisebox{0.2ex}{$\dagger$}}& 94.3 & 47.0 & 60.4 & 45.9& 94.7& 95.8& 95.8 &100.0& 79.2&-\\
     Gemini-2.0 Flash\textcolor{red}{\raisebox{0.2ex}{$\dagger$}}& 52.4&30.6&66.7&31.8&56.0&46.3&24.5&55.1&45.4&-\\
     Gemini-1.5 Pro\textcolor{red}{\raisebox{0.2ex}{$\dagger$}}& 49.6& 28.8& 58.6& 49.4& 46.0& 48.1&  42.0&  68.0& 48.8&-\\
     Gemini-1.5 Pro &  56.2 & 30.9 & 64.1& 43.6& 51.3& 46.3& 36.0& 34.6& 45.4&-\\
     \rowcolor{gray!20}+SpatialMind&63.9&51.8&70.2&47.3&56.3&45.9&42.6&44.3&52.8&\textbf{{\color{darkgreen} $\uparrow$ 7.4\%}}\\
     GPT-4o&46.2 &5.3 &43.8& 38.2 &37.0& 41.3& 31.5& 28.5&34.0&-\\
     \rowcolor{gray!20}+SpatialMind&40.0&27.1&62.7&40.9&41.0&39.6&37.1&38.5&40.8&\textbf{{\color{darkgreen} $\uparrow$ 6.8\%}}\\
     \midrule
    \multicolumn{11}{c}{Open-source}\\
    \midrule
     % DeepSeek& \\
     % +SpatialMind&\\
     InternVL2-8B & 23.1 &28.7& 48.2 &39.8& 36.7 &30.7& 29.9 &39.6& 34.6&-\\
     +SpatialMind&35.8&28.9&49.7&44.4&37.2&34.8&35.1&45.5&38.9&\textbf{{\color{darkgreen} $\uparrow$ 4.3\%}}\\
     
     +ScanForgeQA&45.3&33.4&54.8&45.0&41.1&36.1&33.4&43.0&41.5&\textbf{{\color{darkgreen} $\uparrow$ 6.9\%}}\\
          \rowcolor{gray!20}+Both&47.0&32.8&53.2&46.6&39.8&36.8&37.9&47.5&42.7&\textbf{{\color{darkgreen} $\uparrow$ 8.1\%}}\\
     InternVL2-40B & 34.9& 26.9& 46.5& 31.8 &42.1& 32.2& 34.0 &39.6& 36.0&-\\
     +SpatialMind&36.4&30.0&49.1&41.8&43.8&36.1&35.6&50.0&40.4&\textbf{{\color{darkgreen} $\uparrow$ 4.4\%}}\\
     +ScanForgeQA&51.0&29.2&52.7&38.1&47.2&36.4&35.9&47.6&42.3&\textbf{{\color{darkgreen} $\uparrow$ 6.3\%}}\\
          \rowcolor{gray!20}+Both&52.2&30.5&54.4&41.0&50.5&37.0&40.2&50.3&44.5&\textbf{{\color{darkgreen} $\uparrow$ 8.5\%}}\\
    \midrule
     InternVL2.5-8B&7.0&34.1&43.0&42.4&38.0&40.1&23.2&35.9&33.0&-\\
     +SpatialMind&18.7&29.6&46.5&45.2&40.3&40.9&27.8&46.6&37.0&\textbf{{\color{darkgreen} $\uparrow$ 4.0\%}}\\
     +ScanForgeQA& 56.5&33.2&50.7&43.4&39.0&33.1&28.4&34.3&39.8&\textbf{{\color{darkgreen} $\uparrow$ 6.8\%}}\\
     \rowcolor{gray!20}+Both&52.8&30.7&53.4&44.8&41.1&38.7&29.6&47.0&42.3&\textbf{{\color{darkgreen} $\uparrow$ 9.3\%}}\\
     InternVL2.5-38B&43.6&33.0&53.0&48.8&53.5&35.7&34.5&34.0&42.0&-\\
     \rowcolor{gray!20}+SpatialMind&56.1&38.3&59.3&52.3&59.7&43.6&39.0&32.9&47.7&\textbf{{\color{darkgreen} $\uparrow$ 5.7\%}}\\
     \midrule
     LLaVA-NV-7B & 48.5 &14.0 &47.8& 24.2 &43.5& 42.4 &34.0 &30.6& 35.6&-\\
     \rowcolor{gray!20}+SpatialMind&49.0&22.6&47.9&24.6&41.3&43.0&37.1&29.8&36.9&\textbf{{\color{darkgreen} $\uparrow$ 1.3\%}}\\
          % +ScanForgeQA&\\
          %      \rowcolor{gray!20}+Both&\\
     LLaVA-NV-72B & 48.9 & 22.8 &57.4 &35.3& 42.4 &36.7 &35.0& 48.6&40.9&-\\
     \rowcolor{gray!20}+SpatialMind&52.5&24.5&60.2&37.7&42.7&39.5&37.3&51.0&43.2&\textbf{{\color{darkgreen} $\uparrow$ 2.3\%}}\\
     % +ScanForgeQA&\\
     %      \rowcolor{gray!20}+Both&\\
    \midrule
     % LLaVA-OneVision-72B & 43.5& 23.9 &57.6& 37.5 &42.5& 39.9& 32.5& 44.6& 40.2\\
     VideoLLaMA3-7B&36.8&22.2&34.7&24.9&44.6&41.7&36.1&28.8&33.7&-\\
     \rowcolor{gray!20}+SpatialMind&38.6&23.9&36.2&36.6&42.3&40.9&35.9&33.2&36.0&\textbf{{\color{darkgreen} $\uparrow$ 2.3\%}}\\    
     \midrule
    Qwen2-VL-7B&39.4&25.0&25.8&43.2&32.6&30.9&27.8&32.6&32.2&-\\
    \rowcolor{gray!20}+SpatialMind&43.7&29.0&29.4&46.5&33.9&32.8&32.0&31.9&34.9&\textbf{{\color{darkgreen} $\uparrow$ 2.7\%}}\\
     \midrule

    Qwen2.5-VL-7B&40.3&22.2&50.1&38.9&38.0&40.7&31.4&35.9&37.2&-\\
     +SpatialMind& 45.1&25.2&52.1&41.4&38.7&41.6&34.7&34.5&39.2&\textbf{{\color{darkgreen} $\uparrow$ 2.0\%}}\\
     +ScanForgeQA&53.2&30.5&56.8&44.9&42.3&44.0&37.3&37.7&43.3&\textbf{{\color{darkgreen} $\uparrow$ 6.1\%}}\\
     \rowcolor{gray!20}+Both&55.0&29.5&57.3&44.0&43.5&44.3&38.3&39.2&43.9&\textbf{{\color{darkgreen} $\uparrow$ 6.7\%}}\\
        Qwen2.5-VL-72B&37.9& 28.6&57.4&49.8&45.5&38.4&20.6&35.4&39.2&-\\
     +SpatialMind&42.3&32.0&61.7&53.8&48.2&43.9&30.4&39.3&44.0&\textbf{{\color{darkgreen} $\uparrow$ 4.8\%}}\\
     +ScanForgeQA&45.2&32.7&63.3&52.4&50.1&41.7&32.8&40.2&44.8&\textbf{{\color{darkgreen} $\uparrow$ 5.6\%}}\\
     \rowcolor{gray!20}+Both&48.6&34.4&68.9&54.7&53.4&43.9&30.1&42.7&47.1&\textbf{{\color{darkgreen} $\uparrow$ 7.9\%}}\\
    \bottomrule
  \end{tabular}
    \vspace{-2ex}
  \label{tab:more_model}
\end{table}

\subsubsection{OpenEQA}
OpenEQA~\cite{majumdar2024openeqa}, introduced by Meta AI in 2024, is a benchmark designed to evaluate the spatial understanding and reasoning capabilities of VLMs in real-world indoor environments. The dataset comprises over 1,600 human-annotated question–answer pairs spanning 180+ diverse indoor scenes and supports two primary evaluation tasks. In our experiment, we focused on the \textbf{Episodic Memory} (EM-EQA) setting, where models must answer questions based solely on a previously observed egocentric video trajectory, without access to external spatial priors. Questions cover a wide range of reasoning types, including object attributes, spatial relationships, object locations, and functional reasoning.
To evaluate model perfromance on open-ended responses, we employed GPT-4o~\footnote{\url{https://github.com/mbzuai-oryx/Video-ChatGPT/tree/main/quantitative_evaluation}.} as an automatic evaluator to assess semantic alignment between predicted and ground-truth answers. This enables consistent and reliable scoring across diverse question formats. 

\subsubsection{ScanQA}
ScanQA~\cite{azuma2022scanqa} is a large-scale benchmark designed to evaluate spatial question answering in real-world indoor environments, grounded in richly annotated 3D scans. Built on top of the ScanNet~\cite{dai2017scannet} dataset, it contains over 41,000 human-curated question–answer pairs across 800 RGB-D scenes, with each question grounded in the 3D geometry and semantics of the scene. The dataset covers a broad range of open-ended questions involving object attributes, spatial relations, and scene-level understanding, such as ``What is between the sofa and the table?'' or ``Where is the lamp located?''. To evaluate model performance, we adopted BLEU-1 as the primary metric. Given that the answers are typically short and factual, unigram precision offers a reliable measure of response quality. 

\subsubsection{SQA3D}
SQA3D~\cite{masqa3d} is a reasoning-centric  benchmark designed to evaluate  situational understanding within 3D indoor scenes. Built on 650 scenes from ScanNet~\cite{dai2017scannet}, it comprises approximately 6,800 annotated situations and over 33,000 questions that span a diverse range of reasoning types, including spatial relations, commonsense reasoning, and multi-hop logical reasoning. Each question is grounded in a specific 3D context, requiring precise and spatially informed answers. To evaluate model performance, we used the Exact Match at 1 (EM-1) metric. This metric is well-suited for the benchmark,  as answers are typically concise, specific, and context-dependent.  

\subsection{Baselines} 
As shown in Tables~\ref{tab:more_model}, our evaluation includes a diverse set of VLM baselines spanning a range of architectures and scales.  These include open-source models such as 
InternVL2~\cite{chen2024expanding} (8B, 40B), InternVL2.5~\cite{chen2024expanding} (8B, 38B), LLaVA-NeXT-Video~\cite{li2024llava} (7B, 72B), VideoLLaMA3~\cite{zhang2025videollama} (7B), Qwen2-VL~\cite{wang2024qwen2} (7B), 
and Qwen2.5-VL~\cite{bai2025qwen2} (7B, 72B). We also evaluated closed-source models, including  Gemini-2.0 Flash~\footnote{\url{https://deepmind.google/technologies/gemini/flash/}.}, Gemini-1.5 Pro~\footnote{\url{https://deepmind.google/technologies/gemini/}.} and GPT-4o~\cite{hurst2024gpt}.
For open-source models, we adopted each model's default parameter settings, including learning rate, number of frames, and input resolution. For closed-source models, GPT-4o processes 16 frames per video, while all Gemini models operate at a fixed sampling rate of 1 frame per second (1 FPS). All experiments are conducted on 8 NVIDIA H20 GPUs.

\begin{figure}[t]
  \centering
  \includegraphics[width=0.9\linewidth]{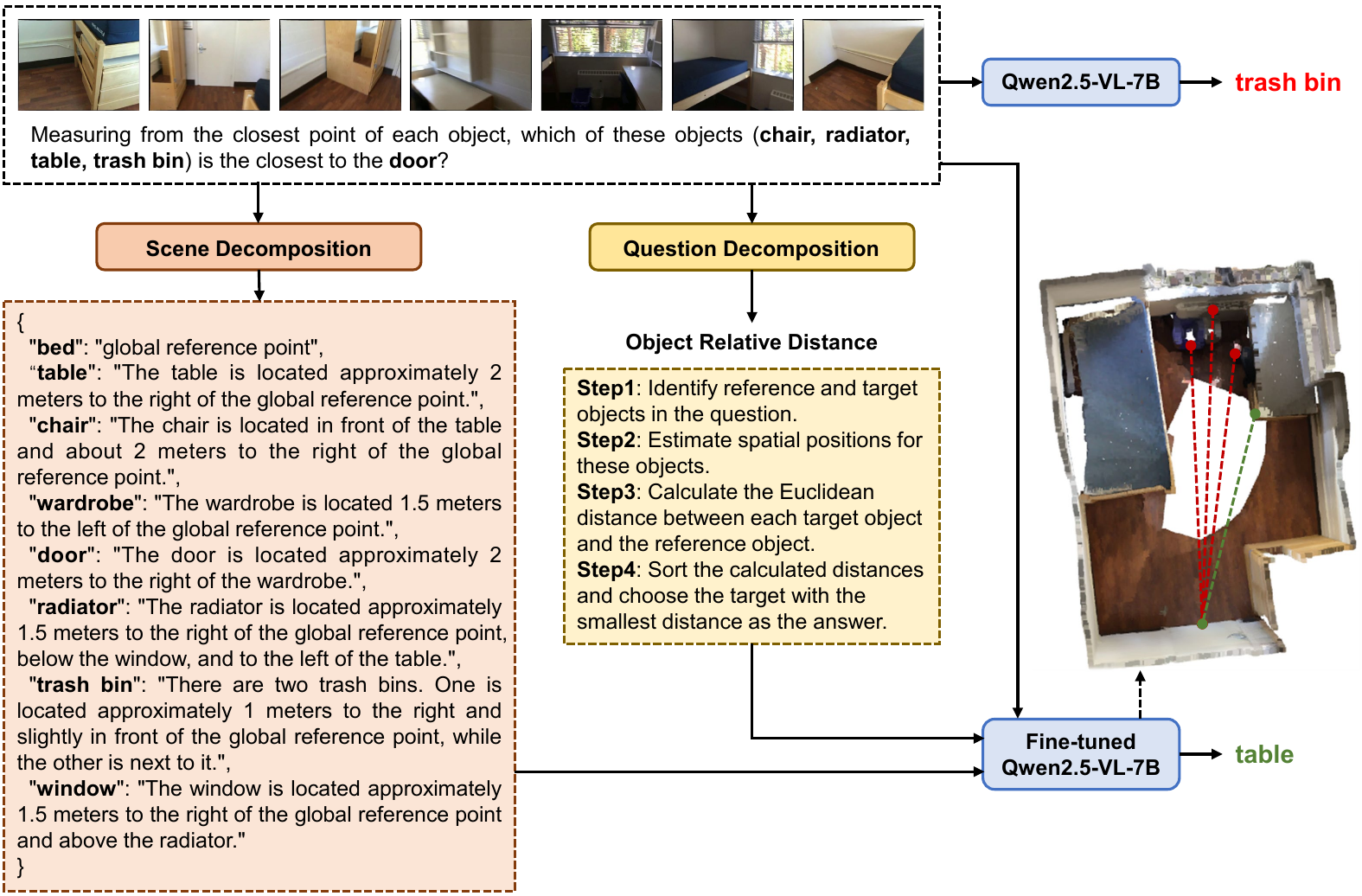}
  \caption{A complete example illustrating the visual prompting process with intermediate outputs.}
  \vspace{-2ex}
  \label{fig:full_case}
\end{figure}

\section{More Experimental Results}\label{app_result}
\subsection{Performance Comparison}
In Table~\ref{tab:more_model}, we extended our evaluation to additional VLMs with varying architectures and parameter scales, applying both the SpatialMind prompting and ScanForgeQA fine-tuning. The results are consistent with those reported in our main analysis, further reinforcing the effectiveness and generalizability of our proposed methods. Moreover, the combined use of prompting and fine-tuning continues to yield superior performance across models, highlighting their complementary strengths and demonstrating the robustness of our framework.
%The results reflect the same trends observed in our main analysis, providing strong evidence for the effectiveness and generalizability of our proposed two methods, as well as the advantages of their combination.

\subsection{Case Study}
Figure~\ref{fig:full_case} presents a detailed example that illustrates the full prompting process along with intermediate outputs. The scene decomposition module produces approximate spatial descriptions of object locations, effectively capturing the overall layout of the 3D scene. Meanwhile, the question decomposition module identifies the question type as ``object relative distance'' and selects the corresponding reasoning steps to guide inference. By combining this structured information and feeding it into the Qwen2.5-VL-7B model fine-tuned with ScanForgeQA, the correct target object is successfully identified.
This example demonstrates the interpretability and effectiveness of our framework in performing multi-step spatial reasoning grounded in visual context.

\section{More details for ScanForgeQA}\label{app_data}
\subsection{Scene Distribution}
\begin{figure}
  \centering
  \includegraphics[width=0.7\linewidth]{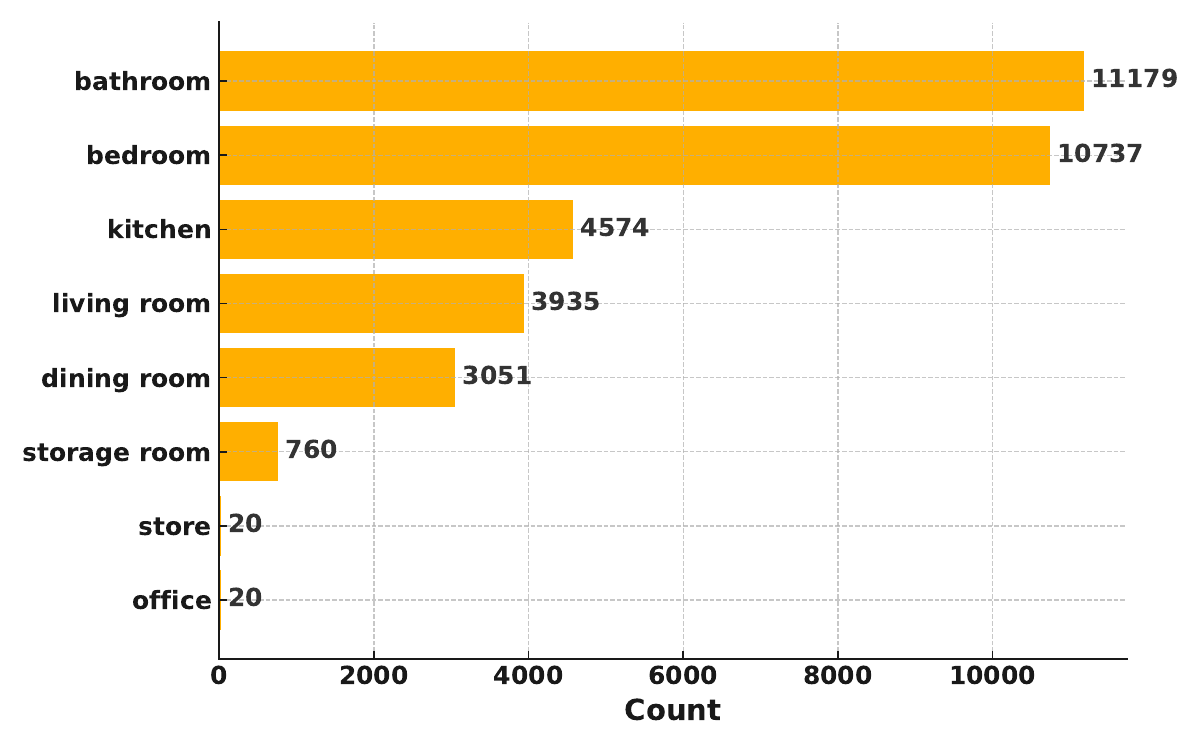}
  \caption{Distribution of room types in the ScanForgeQA dataset.}
  \label{fig:room_dis}
\end{figure}

Figure~\ref{fig:room_dis} presents the distribution of the eight room categories included in ScanForgeQA. The dataset exhibits a clear long-tailed distribution in which categories such as \textit{bathroom} and \textit{bedroom} are heavily represented, whereas \textit{store} and \textit{office} types are comparatively rare. This imbalance is primarily due to the inherent distribution of room types in the 3D-FRONT dataset~\cite{fu20213d}.

\subsection{Scan Example}
\begin{figure}[htbp]
    \centering
    \begin{minipage}[t]{0.49\textwidth}
        \centering
        \includegraphics[width=\linewidth]{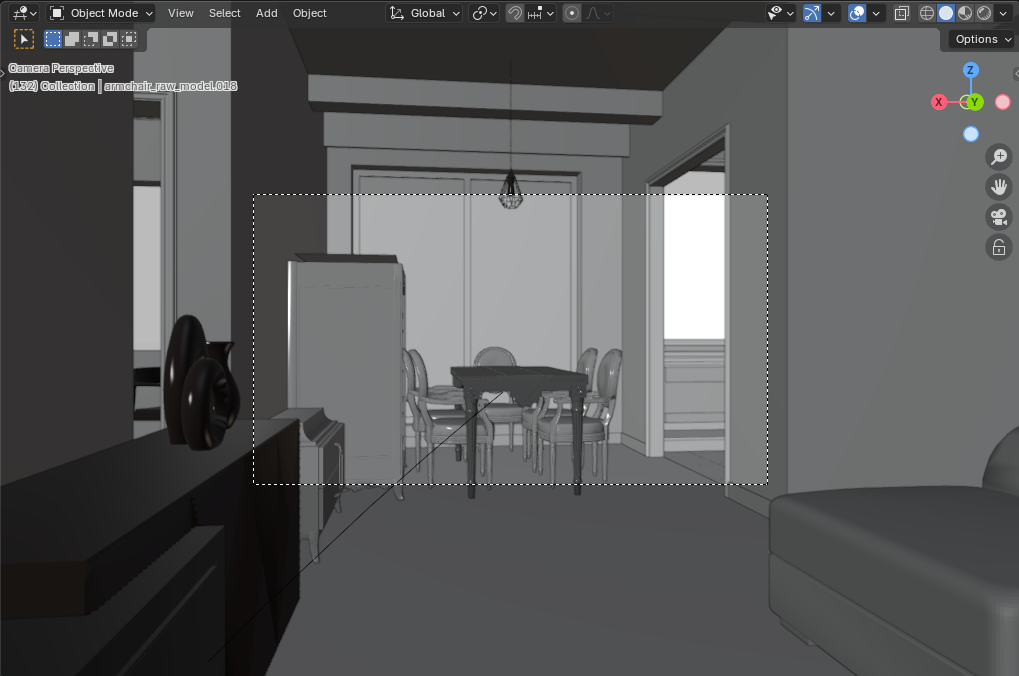}
        \label{fig:image1}
    \end{minipage}
    \hfill
    \begin{minipage}[t]{0.49\textwidth}
        \centering
        \includegraphics[width=\linewidth]{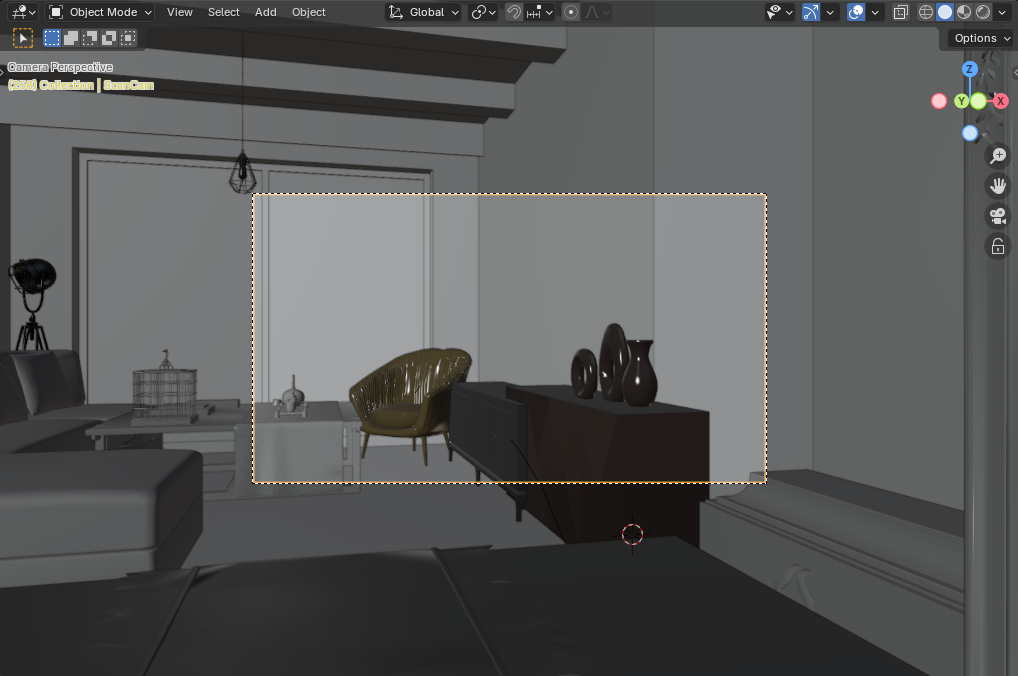}
        \label{fig:image2}
    \end{minipage}
    \caption{Screenshots from the scan creation process.}
    \label{fig:frame}
\end{figure}

We provided two frames captured by the camera during the creation of the scanning video, as shown in Figure~\ref{fig:frame}, to visually illustrate this process. Additionally, we further provided scanning creation scripts as well as more video demos in the Supplemental Material:
\begin{itemize}
    \item ``\textbf{codes/nav\_script.py}'' is the script that creates the navigation scan, presenting the exact implementation.
    \item ``\textbf{Creation of Scanning Video.mp4}'' illustrates the creation process of the scanning video from both first-person and third-person perspectives.
    \item ``\textbf{Scanning Video.mkv}'' presents the rendered scene scan resulting from the aforementioned scanning process.
\end{itemize}

\subsection{QA Definition}
%We provide templates and answer sources for each type of question, as shown in Table~\ref{tab:question-types}. These details clearly illustrate which question templates are covered by each type, as well as which scene annotations the corresponding answers are based on.
Table~\ref{tab:question-types} provides representative templates and corresponding answer sources for each question type in ScanForgeQA. This information clarifies the coverage of different reasoning categories and indicates which scene annotations are used to derive ground-truth answers. 

\begin{table}[t]
\centering
\caption{Overview of question templates and their corresponding answer sources.}
\begin{tabularx}{\linewidth}{lp{6cm}Y}
\toprule
\textbf{Type} & \textbf{Question Template} & \textbf{Answer Source} \\
\midrule
\multicolumn{3}{c}{\textbf{Attribute Estimation}}\\
\midrule

\multirow{2}{*}{Object Count} & How many <A> are there in the room? & \multirow{2}{=}{Name of object instance} \\
& What is the total number of <A>?&\\

\multirow{6}{*}{Object Size} & What is the length of the longest side of <A> in meters? & \multirow{6}{=}{Dimension of the object model} \\
&What is the size of <A> in square meters?&\\
&What is the length of the shortest side of <A> in meters?&\\
&How tall is <A> in meters?&\\

\multirow{2}{*}{Room Size} & What is the size of the room in square meters? & Scene boundaries and scale \\

\multirow{4}{*}{Room Type} & Based on object layout, what is the most likely type of this room? & \multirow{4}{*}{Room labels} \\

&Is this space a living room, a kitchen, or something else?&\\

\midrule
\multicolumn{3}{c}{\textbf{Spatial Reasoning}}\\
\midrule
\multirow{4}{*}{Relative Distance} & Which of these objects (<A>, <B>, <C>) is the closest to <R>? & \multirow{4}{=}{The minimum distance between the centroids of objects} \\

&Among the listed objects (<A>, <B>, <C>), which one is closest to <R>?&\\

\multirow{5}{*}{Absolute Distance} & What is the distance between <A> and <B> in meters? & \multirow{5}{=}{Euclidean distance between objects} \\
&Measure the distance from <A> to <B> in meters.&\\
&How far is <A> from <B> in meters?&\\

\multirow{4}{*}{Relative Direction} & If I am standing by <A> and facing <B>, which side is object <R> on? & \multirow{4}{=}{The direction of connections between objects}\\

&From the viewpoint at <A> facing <B>, where is <R>?&\\

\multirow{2}{*}{Contact Relationship} & Is there a gap between <A> and <B>? & \multirow{2}{=}{Comparison of distances and sizes between objects} \\
&Are <A> and <B> touching each other?&\\

\midrule
\multicolumn{3}{c}{\textbf{Hypothesis Analysis}}\\
\midrule
\multirow{4}{*}{Operation Feasibility} & Considering only object sizes, is there enough space to put <A> in <B>? &  \multirow{4}{=}{Comparison of sizes between objects}\\

&Considering only object dimensions, is it feasible to place <A> on <B>?&\\

\bottomrule
\end{tabularx}
\label{tab:question-types}
\end{table}

\section{More details for SpatialMind}\label{app_cot}
\subsection{Prompt for Object Description}
As a representative example, Figure~\ref{fig:prompt-example} illustrates the prompt used to generate textual descriptions of object positions. This prompt guides the model to produce structured spatial layouts in natural language based on the visual input. Similar prompts are used for other reasoning components, following a consistent design pattern. Other prompts are provided in the ``\textbf{codes/gen\_scene\_exp.py}'' file of Supplementary Material for reference and reproducibility.

\subsection{Reasoning Steps for Different Questions}
Figure~\ref{fig:full_case} provides an example of the detailed reasoning steps used for the \textit{relative distance} question type. Each question type is paired with a concise, structured reasoning process that offers a generalizable solution strategy for VLMs. All reasoning steps for the various question types are included in the ``\textbf{codes/reason\_steps.py}'' file of Supplementary Material for reference and reproducibility.

\section{Limitations}\label{limitation}
This paper demonstrates that existing VLMs tend to rely more heavily on understanding textual descriptions when performing spatial reasoning. However, textual descriptions alone may not capture spatial semantics as intuitively as visual input, potentially limiting the upper bound of performance for such methods. Moreover, we primarily focus on basic single-room 3D scene. Although strong performance has been achieved on OpenEQA~\cite{majumdar2024openeqa}, which includes some multi-room scenes, more complex settings, such as large-scale indoor layouts and outdoor environments, remain underexplored and represent important directions for future work.

\section{Broader Impacts}\label{impact}
Our approach improves visual-spatial understanding in VLMs, enabling better interpretation of object layouts and spatial relationships. This enhancement offers potential benefits in areas such as embodied AI and AR/VR applications, where spatial reasoning is essential. However, similar to other prompting and fine-tuning methods, our framework may inherit biases from pre-trained models or training data, potentially leading to unfair or inaccurate spatial interpretations. While our method reduces reliance on expensive 3D sensors, evaluating spatial accuracy remains a key challenge, especially for downstream tasks that demand high robustness and trustworthiness.  Despite these limitations, we believe that releasing our framework and dataset will foster transparency, support reproducibility, and accelerate progress toward trustworthy and robust spatial reasoning systems.

\section{More Related Work}
\textbf{Vision Language Models} (VLMs) are multimodal systems that combine visual perception with natural language understanding, enabling them to process and generate content that integrates both images and text. Their early development is driven by pretraining on large-scale image-text pairs, which laid the foundation for integrating these models with  LLMs~\cite{schuhmann2022laion,yang2024vision,guan2022bi}.
Initial approaches typically use attention mechanisms or intermediary modules such as Q-Formers to fuse visual and textual information before feeding it into LLMs~\cite{alayrac2022flamingo,huang2023language,li2022blip,li2023blip}. A more efficient approach later emerges: directly mapping image features into the LLM's input space using multilayer perceptrons (MLPs), which led to notable performance and simplicity~\cite{liu2023visual,zhu2023minigpt,liu2024improved,wang2024qwen2}. As the field progressed, VLMs expand beyond basic image-text alignment to support tasks like visual grounding~\cite{chen2023minigpt,lai2024lisa,sunemu,wang2024cogvlm,xiao2024can,liu2018attentive,liu2018cross}. They are later adapted for video understanding, where spatiotemporal compression techniques allow models to handle long sequences of visual data efficiently~\cite{cheng2024videollama,zhang2025exo2ego,qian2024streaming,chen2024sharegpt4video,xing2024pyramiddrop,zhang2024sparsevlm,wan2024look,wang2025time}.

While some recent studies have begun to investigate the application of VLMs to indoor scene understanding, most existing efforts remain constrained to standardized benchmarks~\cite{yang2024thinking,li2025sti} or continue to depend on traditional 3D-based methods~\cite{zhang2024agent3d,zheng2024video}. In this work, we introduce two novel approaches that empower VLMs to perceive and reason about the 3D world solely from visual inputs, marking a significant advancement in their capabilities.

\vspace{2ex}
\begin{tcolorbox}[title=Prompt, promptbox]
[Task]\\
You are given a video (multiple frames) capturing an indoor scene. Your goal is to recognize \{categories\_of\_interest\} objects, analyze the spatial layout of the scene, and describe the relative position of each object.\\

[Instructions]\\
1. Per-frame analysis:\\
- For each frame, choose one object as a **local reference point**.\\
- Predict the relative position of all other objects with respect to this local reference point.\\
- Express relative positions using simple terms like "left", "right", "front", "behind", "above", "below", and approximate distances (e.g., "2 meters to the right").\\

2. Global scene layout:\\
- Take the **local reference point from the first frame** as the **global reference point** for the whole video.\\
- Use overlapping objects between frames to align frames together.\\
- Gradually build the spatial descriptions for all objects relative to the global reference point.\\

[Rules]\\
- If a category has multiple instances (e.g., two chairs), describe each instance separately.\\
- Preserve the real-world spatial relationships and distances as accurately as possible.\\
- Use clear and consistent directional and distance terms.\\

[Output Format]\\
ONLY Return the result as a JSON dictionary following STRICTLY this format:\\
\{\\
    "category name": "global reference point",\\
    "another category name": "Position description relative to the global reference point",\\
    ...\\
\}\\

Example:\\
\{\\
    "chair": "The chair is located 1.5 meters to the left and 0.5 meters behind the global reference point",\\
    "table": "The table is located 2 meters to the right of the global reference point",\\
    "lamp": "The lamp is located 1 meter above and 1 meter behind the global reference point"\\
\}
\end{tcolorbox}
\captionof{figure}{Example prompt used to generate textual descriptions of object positions.}\label{fig:prompt-example}

\end{document}